**Advanced Hybrid Transformer–LSTM Technique with Attention and TS-Mixer for Drilling Rate of Penetration Prediction**

Saddam Hussain Khan[1]*
[1]Artificial Intelligence Lab, Department of Computer Systems Engineering, University of Engineering and Applied Sciences (UEAS), Swat, Pakistan
**Email:** saddamhkhan@ueas.edu.pk

## Abstract

Rate of Penetration (ROP) prediction is critical for drilling optimization yet remains challenging due to the nonlinear, dynamic, and heterogeneous characteristics of drilling data. Conventional empirical, physics-based, and standard machine learning models rely on oversimplified assumptions or intensive feature engineering, constraining their capacity to model long-term dependencies and intricate feature interactions. To address these issues, this study presents a new deep learning 'Hybrid LSTM-Trans-Mixer-Att' framework that first processes input data through a customized Long Short-Term Memory (LSTM) network to capture multi-scale temporal dependencies aligned with drilling cycles. Subsequently, an Enhanced Transformer encoder with drilling-specific positional encodings and real-time optimization refines the features. Concurrently, a parallel Time-Series Mixer (TS-Mixer) block introduced facilitates efficient cross-feature interaction modeling of static and categorical parameters, including lithological indices and mud properties. The feature representations extracted from the Enhanced Transformer and TS-Mixer modules are integrated through a dedicated fusion layer. Finally, an adaptive attention mechanism then dynamically assigns contextual weights to salient features, enhancing discriminative representation learning and enabling high-fidelity ROP prediction. The proposed framework combines sequential memory, static feature interactions, global context learning, and dynamic feature weighting, providing a comprehensive solution for the heterogeneous and event-driven nature of drilling dynamics. Experimental validation on real-world drilling datasets demonstrates superior performance, achieving an $R^2$ of 0.9991 and a MAPE of 1.447%, significantly outperforming existing baseline and hybrid models. Furthermore, interpretability analyses using SHAP and LIME confirm the model's transparency, fairness, and robustness, establishing it as a viable solution for intelligent and cost-efficient drilling optimization systems.

**Keywords**: Rate of Penetration (ROP), Deep Learning, Transformer, LSTM, Time Series (TS)-Mixer, Drilling, Optimization, Attention, Forecasting.

# 1 Introduction

The Rate of Penetration (ROP) is a crucial performance indicator in drilling operations, directly impacting drilling efficiency, costs, and project schedules. Accurate and timely ROP forecasting is crucial for making informed decisions, optimizing drilling in real-time, reducing risks, and minimizing non-productive time [1]. Traditional ROP prediction methods, such as empirical correlations, expert heuristics, and physics-based models, often lack flexibility in complex and changing subsurface environments [2]. These methods are limited because they cannot effectively capture the nonlinear relationships and high-dimensional interactions among geological, mechanical, and operational factors that influence drilling performance.

Recent advances in artificial intelligence (AI) and machine learning (ML) have introduced powerful data-driven alternatives capable of addressing these limitations. By learning intricate, nonlinear patterns from large volumes of drilling data, AI-based approaches provide new opportunities for developing intelligent drilling optimization frameworks [3]. In particular, deep learning (DL) architectures such as Long Short-Term Memory (LSTM) networks and Transformer models [4]. Moreover, feature-mixing frameworks like TS-Mixer exhibit strong capabilities in extracting sequential dependencies, modeling long-range correlations, and learning high-order feature interactions [5]. These models have achieved promising results in predicting ROP, yet challenges persist in ensuring robust generalization across diverse drilling environments and operational contexts.

To address these limitations, hybrid DL strategies have emerged as a promising approach. By combining the sequential memory capabilities of LSTM networks and the attention mechanisms of Transformers [6]. The dense feature interactions of TS-Mixers, hybrid models, can more effectively capture temporal dynamics, contextual dependencies, and cross-feature relationships [7]. This integration allows for a more comprehensive representation of drilling processes, resulting in improved prediction accuracy and greater resilience across diverse drilling conditions. Consequently, this study investigates a hybrid DL framework for ROP prediction, aiming to enhance intelligent drilling systems that improve efficiency, safety, and sustainability in oil and gas exploration. Hybrid models provide several key advantages in the context of ROP prediction.

1. Sequential dependency modeling through LSTM layers that learn temporal correlations across drilling cycles, reflecting the effects of operational variations over time.
2. Cross-feature interaction learning via fully connected TS-Mixer modules that effectively merge static and dynamic drilling parameters.
3. Global contextual attention through Transformer encoders which emphasize the most salient features irrespective of their spatial or temporal positions.
4. Error reduction and robustness are enhanced by combining multiple architectural perspectives, improving generalization, and mitigating overfitting crucial for error-sensitive drilling applications [8].

ROP prediction has primarily utilized sequential models, such as LSTM and GRU, as well as attention-based Transformers [9]. Yet, these approaches often suffer from limited generalization and weak integration of multi-dimensional drilling features. To overcome these challenges, this work proposes a hybrid DL framework that integrates a customized LSTM, an Enhanced TS-Mixer [10], Transformer encoders, and attention mechanisms [11] [12]. Moreover, the framework demonstrated computational efficiency, ensuring feasibility for real-time deployment in drilling operations. The systematic experimental analysis underscores the transformative potential of hybrid DL architectures in advancing drill performance prediction and operational decision-making. The main contributions of this work can be summarized as follows:

- A novel hybrid ‘Hybrid LSTM-Trans-Mixer-Att’ DL framework is developed that integrates a customized LSTM, an Enhanced Transformer encoder, a Time-Series Mixer (TS-Mixer), and an adaptive attention mechanism to jointly capture temporal dependencies, cross-feature interactions, and global contextual patterns for dynamic feature prioritization.
- The customized LSTM with cycle-aligned encoding and multi-scale stacking models hierarchical temporal dependencies and cyclic drilling dynamics, while its output feeds into an Enhanced Transformer with domain-specific positional encodings for real-time long-range context learning. Concurrently, the TS-Mixer processes the original input to capture nonlinear interactions among static and categorical drilling parameters, enhancing representational depth and cross-domain coherence.
- An adaptive attention fusion mechanism then unifies outputs from the sequential (customized LSTM–Enhanced Transformer) and static (TS-Mixer) streams through dual-level contextual weighting, emphasizing salient temporal–spatial dependencies, reducing redundancy, and improving discriminative power. Collectively, the proposed framework enhances prediction accuracy, interpretability, and robustness, offering a scalable foundation for intelligent drilling optimization.
- Extensive testing on real-world drilling datasets shows the framework’s ability to model nonlinear and diverse drilling dynamics, consistently surpassing standalone models, earlier hybrids, and traditional regression methods. These results confirm the framework’s robustness and adaptability across various drilling conditions while delivering top performance with an $R^2$ of 0.9988 and a MAPE of 1.447%.
- The proposed Framework transparency, fairness, and adaptability are maintained through the integration of SHAP- and LIME-based interpretability, bias analysis, and a modular design. The superior performance across diverse evaluation metrics substantiates its potential for real-time deployment. Moreover, it supports trustworthy decision-making, as well as easy extension to other drilling performance indicators and time-series forecasting domains.

The proposed Framework transparency, fairness, and adaptability are maintained through the integration of SHAP- and LIME-based interpretability, bias analysis, and a modular design. This not only confirms the accuracy and operational readiness of the framework for real-time deployment but also supports trustworthy decision-making and easy extension to other drilling performance indicators and time-series forecasting domains.

The remainder of this paper is structured as follows. Section 2 reviews related work on ROP prediction and hybrid DL approaches. Section 3 details the proposed methodology, including data preprocessing, model architecture, and training strategies. Section 4 presents experimental results along with comparative analyses against baseline and hybrid models. Section 5 discusses the key findings and their implications for drilling engineering practice. Finally, Section 6 concludes the study and outlines potential directions for future research.

## 2 Related Work

ROP prediction is a long-standing challenge in drilling engineering. Early studies relied on empirical correlations and physics-based models with limited adaptability under dynamic downhole conditions [13]. Subsequent work applied classical ML to better capture multi-parameter interactions, while recent advances in DL have enabled more effective modeling of nonlinear and temporal dependencies [2], [14]. Hybrid DL approaches, in particular, have shown promise by integrating complementary architectures. This section reviews prior studies in three categories: (1) empirical and physics-based models, (2) classical ML methods, and (3) DL approaches, with emphasis on hybrid DL frameworks, while identifying limitations that motivate this work.

## 2.1 Conventional Empirical Models

Early efforts in ROP prediction were based on empirical relations and physics-based formulations, such as the Bingham model [2]. These approaches expressed ROP as a function of drilling parameters, including weight on bit (WOB), rotary speed (RPM), bit type, mud properties, and formation strength [15]. Although foundational and still applied in practice, they rely on oversimplified assumptions of linearity, homogeneity, and constant operating conditions, which limit accuracy under heterogeneous and dynamic geological settings [16].

Physics-based models introduced rock failure mechanics and bit–rock interaction effects, yet their effectiveness remains restricted by site-specific calibration requirements and the unavailability of accurate real-time drilling data [17]. Their dependence on complex calibration and high-fidelity datasets also hinders scalability across diverse drilling environments. These limitations underscore the transition toward data-driven methods.

## 2.2 Machine Learning and Deep Learning Based ROP Prediction

ML and DL approaches address nonlinearities, adapt to heterogeneous data, and exploit real-time streams, offering more generalizable and operationally robust solutions for ROP prediction. ML Methods such as SVR, Random Forests (RF), and XGBoost capture nonlinear interactions and high-dimensional feature spaces with improved accuracy [18], while K-Nearest Neighbors (KNN) and ensemble methods further enhance short-term forecasts [8, 34]. However, their reliance on limited sequential modeling capacity and heavy feature engineering constrains scalability [19].

DL methods overcome these limitations by learning hierarchical features directly from sequential data. Recurrent Neural Networks (RNNs) and LSTM architectures effectively capture long-range temporal dependencies [20]. Nonetheless, LSTMs remain sensitive to hyperparameters, noise, and data imbalance, restricting robustness in field-scale deployment. Transformers address long-range dependency modeling via self-attention, eliminating recurrence and enabling efficient training [21]. While promising, their use in drilling applications alone or in hybridization with LSTMs remains underexplored [15]. Concurrently, feature-mixing networks such as TS-Mixer provide efficient alternatives by modeling feature interactions through channel and feature mixing; however, their application to drilling, particularly within hybrid frameworks, is largely untapped [5].

Recent studies have incorporated attention mechanisms to dynamically weight features or time steps, enhancing interpretability and emphasizing critical drilling factors [22]. Yet, these are often employed in isolation rather than integrated within architectures that unify sequential memory (LSTM), global context learning (Transformers), and feature interaction modeling (TS-Mixer) [40]. This motivates the present work, which systematically bridges these paradigms into a hybrid framework optimized for ROP prediction. A comparative summary of these prior approaches and their reported limitations is presented in Table 1.

## 2.3 Deep Hybrid Learning

The deep hybrid learning architectures, which integrate multiple paradigms, have recently demonstrated superior predictive performance across domains such as healthcare, finance, and autonomous systems [23], [24], [25], [26], [27]. By combining complementary modeling capabilities, temporal memory (LSTM), global context learning (Transformers), and nonlinear feature interactions (fully connected networks), these approaches significantly reduce prediction error and enhance generalizability [28], [29].

In drilling, however, the adoption of hybrid models remains limited. Existing studies have primarily explored concatenations of LSTM and CNN layers, or GRU networks augmented with attention mechanisms for ROP prediction [9][30]. Yet, critical components such as sequential memory via LSTM, feature interaction learning through TS-Mixer [5], global dependency modeling using Transformers, and dynamic feature weighting with attention mechanisms have rarely been unified within a single architecture.

Table 1. ROP Prediction Studies.

| Study | Dataset | Methods | Results | Key Advantages | Challenges |
|---|---|---|---|---|---|
| Enhanced Version [31] (2022) | Niger Delta (1000–9000 ft): WOB, RPM, mud weight, standpipe pressure, torque, flow rate | Modified Bourgoyne & Young Model with multiple linear regression, stepwise optimization | $R^2$ = 0.972; error cut from 0.03% to 0.003% post-optimization | 2-parameter model; high accuracy in floundering regions; field-ready. | Niger Delta–specific; retraining required elsewhere; assumes constant conditions. |
| ANN 2022, [32]) | Three vertical onshore wells (4525 pts): torque, flow rate, RPM, WOB, standpipe pressure, UCS | ANN with optimized weights and biases, compared with 4 existing ROP models | R = 0.94; AAPE = 8.6%; surpassed Maurer (0.72), Bingham (0.87), B&Y (0.89). | Outperforms traditional models; ANN-derived empirical equation. | Only for carbonate formations and vertical wells; needs an identical formation type. |
| BiLSTM-SA-IDBO (2024) [14]) | Oil drilling with the Bingham equation: WOB, RPM, drilling fluid parameters, and formation properties | BiLSTM + Self-Attention + Improved Dung Beetle Optimization | RMSE = 0.065; $R^2$ = 0.963; MAE = 0.05; 78% RMSE improvement over BiLSTM-SA. | Physical equations; bidirectional temporal processing; robust optimization | Relies on teacher model quality, complex tuning, and high computational cost |
| LSTM Prediction (2025) [33] | Offshore drilling: WOB, RPM, drilling parameters, multi-well formation data | LSTM with particle swarm optimization for hyperparameter tuning | $R^2$ = 0.978; RMSE = 0.287; MAPE = 12.86%; 44.2% accuracy boost. | Captures long-term dependencies; handles temporal correlations. | Hyperparameter- and noise-sensitive; overfits without regularization |
| PCA-Informer Model (2024) [34] | Taipei Basin: depth, gamma, formation density, pore pressure, well diameter, drilling time, displacement | PCA + Informer NN for long-term sequence prediction | MAE = 9.402; RMSE = 0.172; $R^2$ = 0.858; outperformed RNN and LSTM. | Ultra-long ROP prediction; dimensionality reduction; efficient high-computation. | Basin-specific; weaker long-range modeling than pure Transformers |
| CBT-LSTM (2024) [35] | Chinese oilfield (4 vertical wells): drilling parameters, SG filter, PCA reduction | 2D-CNN + BiLSTM + Temporal Pattern Attention (TPA) mechanism for feature extraction and prediction | MAE = 0.0295; MAPE = 0.0357; RMSE = 9.31%; $R^2$ = 0.9769; generalization $R^2$ > 0.95. | Multi-feature extraction; bidirectional processing; noise-resistant; generalizable. | High complexity; heavy preprocessing; sensitive to data quality and noise. |

## 2.4 Literature Gaps Review

Continuum-based theoretical models like the Discrete Element Method and Finite Element Method have long been applied to study fluid-induced forces in porous media. However, within the context of ROP prediction, several critical gaps persist:

- Overreliance on single-architecture models (e.g., pure LSTM or Transformer), with limited exploitation of hybridization benefits.
- Insufficient simultaneous modeling of local feature-wise interactions and global sequence-wise dependencies.
- Suboptimal integration of attention mechanisms within feature-mixing frameworks [36].
- Lack of systematic comparative studies on hybrid architectures using standardized drilling performance metrics ($R^2$, MAE, RMSE, MAPE) applied to real field datasets [16].

However, critical components like sequential memory with LSTM, feature interaction through TS-Mixer, global dependency modeling using Transformers, and dynamic feature weighting with attention mechanisms have rarely been combined within a single framework.

## 3 Methodology

This study introduces a new hybrid DL "Hybrid LSTM-Trans-Mixer-Att" framework that integrates sequential memory, global context modeling, feature interaction learning, and adaptive attention, offering a state-of-the-art solution for ROP prediction. The proposed methodology follows a structured workflow comprising data acquisition, preprocessing, feature encoding, and DL-based "Hybrid LSTM-Trans-Mixer-Att" framework for ROP prediction (Figure 1). Raw drilling parameters are standardized and temporally aligned to maintain consistency across varying operational conditions. The preprocessed sequences are processed through a customized LSTM to capture hierarchical temporal dependencies inherent in drilling dynamics. The resulting embeddings are refined using an Enhanced Transformer encoder with drilling-specific positional encodings and optimized self-attention for modeling long-range dependencies. In parallel, a Time-Series Mixer (TS-Mixer) module captures nonlinear cross-feature interactions among categorical and static parameters, including lithology, mud properties, and bit configurations.

Feature representations from the enhanced Transformer and TS-Mixer are fused through an adaptive attention mechanism. This concurrent use of LSTM-based sequential memory, TS-Mixer-driven feature interaction, Transformer-based global dependency modeling, and dynamic attention weighting within a unified architecture remains largely unexplored. The mechanism dynamically highlights the most informative features, ensuring robust and precise ROP estimation across diverse geological settings. Model interpretability is achieved through SHAP and LIME analyses, which generate post-hoc explanations linking influential inputs to drilling performance. The integration of these components enhances model transparency, trustworthiness, and operational relevance [37]. The methodology is organized into successive stages, each elaborated in the following subsections.

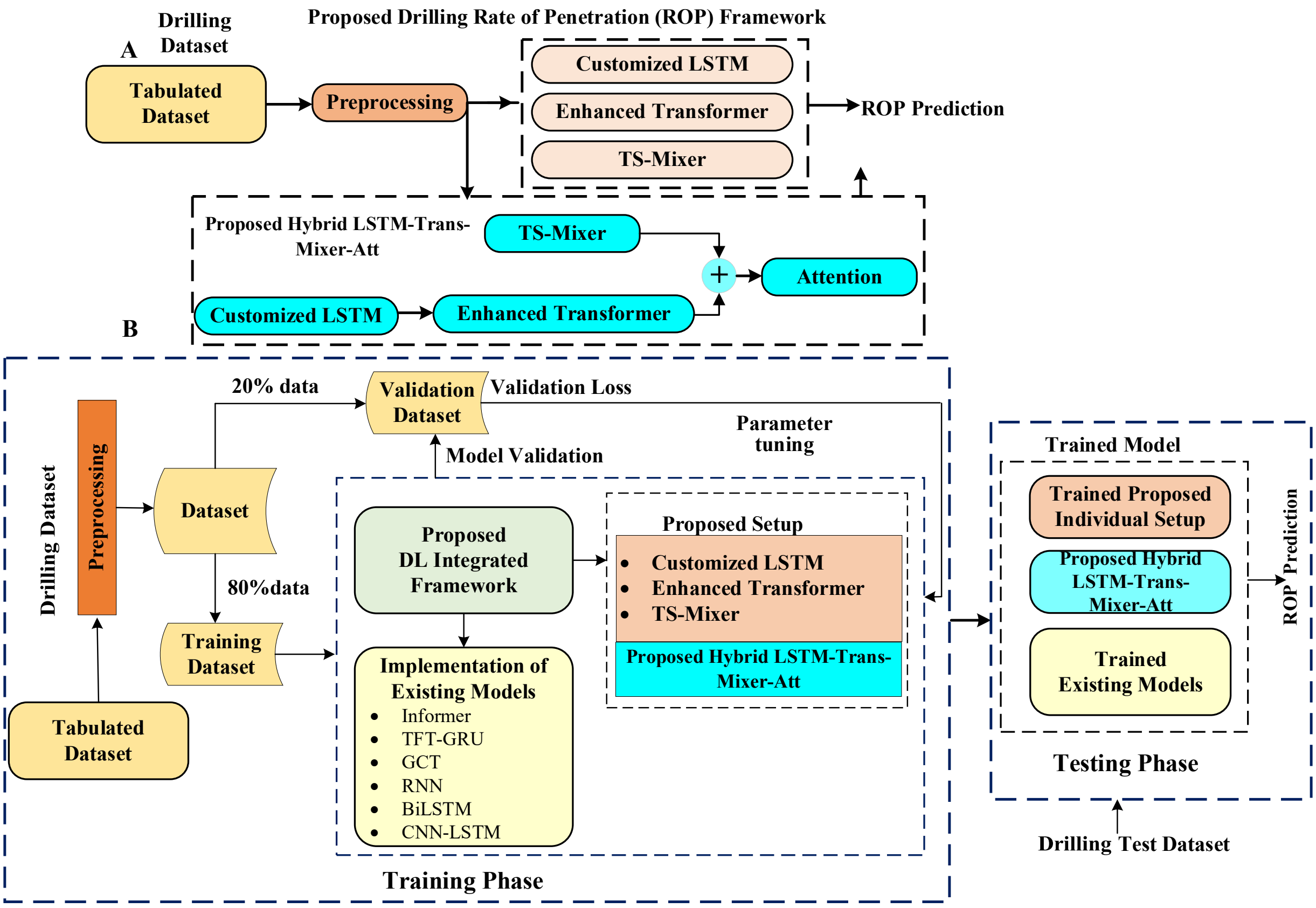

Figure 1: Overall methodology of the proposed drilling ROP Prediction.

## 3.1 Dataset Description

This study utilizes an open-source drilling dataset from Well-1 in the Norwegian oil fields, which is publicly available on GitHub. The dataset captures high-resolution, jack-by-jack drilling records across multiple geological formations, thereby resolving the ROP response to rapid operational fluctuations. Both surface and downhole parameters are included, enabling a holistic representation of drilling dynamics. The recorded features comprise standard operational parameters such as WOB, Rotary Speed (RPM), Standpipe Pressure, Surface Torque, and lithological descriptors, as summarized in Table 2. The target variable, ROP (m/h), is continuously logged across drilling intervals. Moreover, the correlation structure among surface and downhole features is shown in Figure 2, revealing strong interdependencies essential for accurate ROP prediction.

The dataset comprises approximately 10,672 samples spanning the entire drilling sequence, from surface casing through intermediate sections to the production zone. This provides broad geological and operational diversity, including variations in lithology, mud type, pressure regimes, and bit design. Such heterogeneity ensures the dataset is well-suited for developing and validating generalized predictive models, with strong applicability across diverse drilling environments.

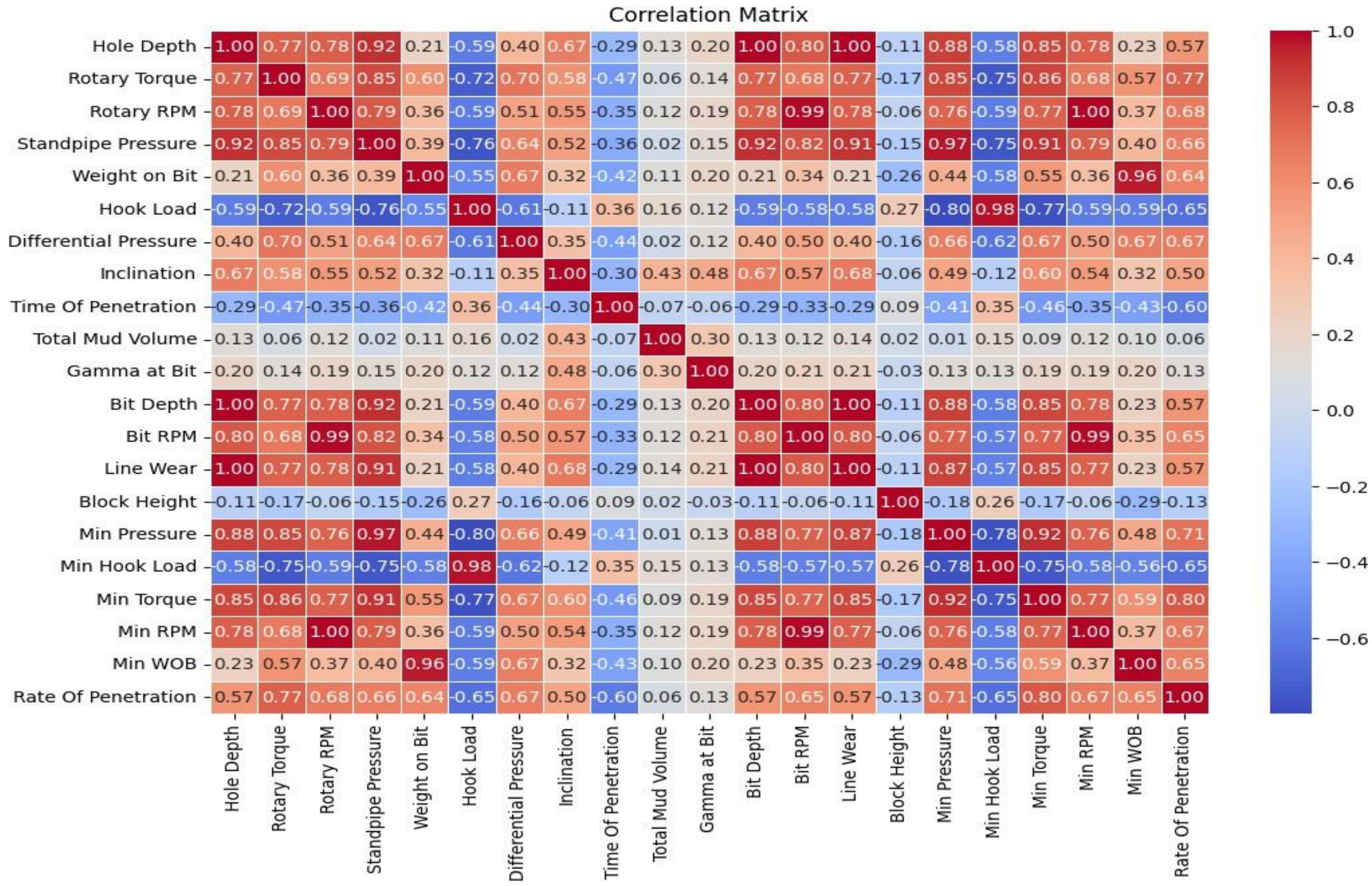

Figure 2: Heatmap of Pearson correlation coefficients among surface and downhole features in the Well-1 dataset, illustrating strong interdependencies crucial for accurate ROP prediction.

Table 2. Features of the Drilling Dataset.

| Feature Name | **Description** | **Unit** | **Type** |
|---|---|---|---|
| WOB | Force applied to the drill bit | klbf | Continuous |
| RPM | Rotational speed of the drill bit | revolutions/min | Continuous |
| Torque | Torque applied during drilling | klbf-ft | Continuous |
| Standpipe Pressure | Pressure within the standpipe | psi | Continuous |
| Flow Rate | Mud flow rate through the drill pipe | gallons/min | Continuous |
| Hook Load | Total weight carried by the hook | klbf | Continuous |
| Bit Depth | Depth reached by the drill bit | feet | Continuous |
| Hole Depth | Total depth of the borehole | feet | Continuous |
| ROP | Target variable drilling progress rate | feet/hour | Target |

## 3.2 Data Preprocessing

Data preprocessing is critical for ensuring reliability in industrial drilling datasets, which are often noisy, incomplete, and affected by sensor or manual errors. A structured multi-step pipeline (Table 3) has been implemented to preserve data integrity. Initial QA/QC analysis revealed missing entries, primarily in infrequently recorded parameters such as Bottom Hole Assembly (BHA) settings and lithology points. Numerical attributes with less than 5% missingness were treated using mean imputation to minimize distributional distortion, while categorical variables were assigned an "Unknown" label to retain their structural integrity and prevent artificial bias. No feature exceeded 15% missing data, eliminating the need for feature removal due to excessive information loss [38]. Moreover, the distributions of key drilling features (Figure 3) highlight data normalization and variability.

Outliers were identified using the Interquartile Range (IQR) method, flagging values beyond 1.5×IQR from the lower or upper quartiles (Figure 4). Expert review confirmed that extreme cases, such as very low ROP in hard formations, were operationally valid. Therefore, no values were removed, preserving the dataset's full operational envelope and enabling models to generalize across both typical and extreme drilling regimes.

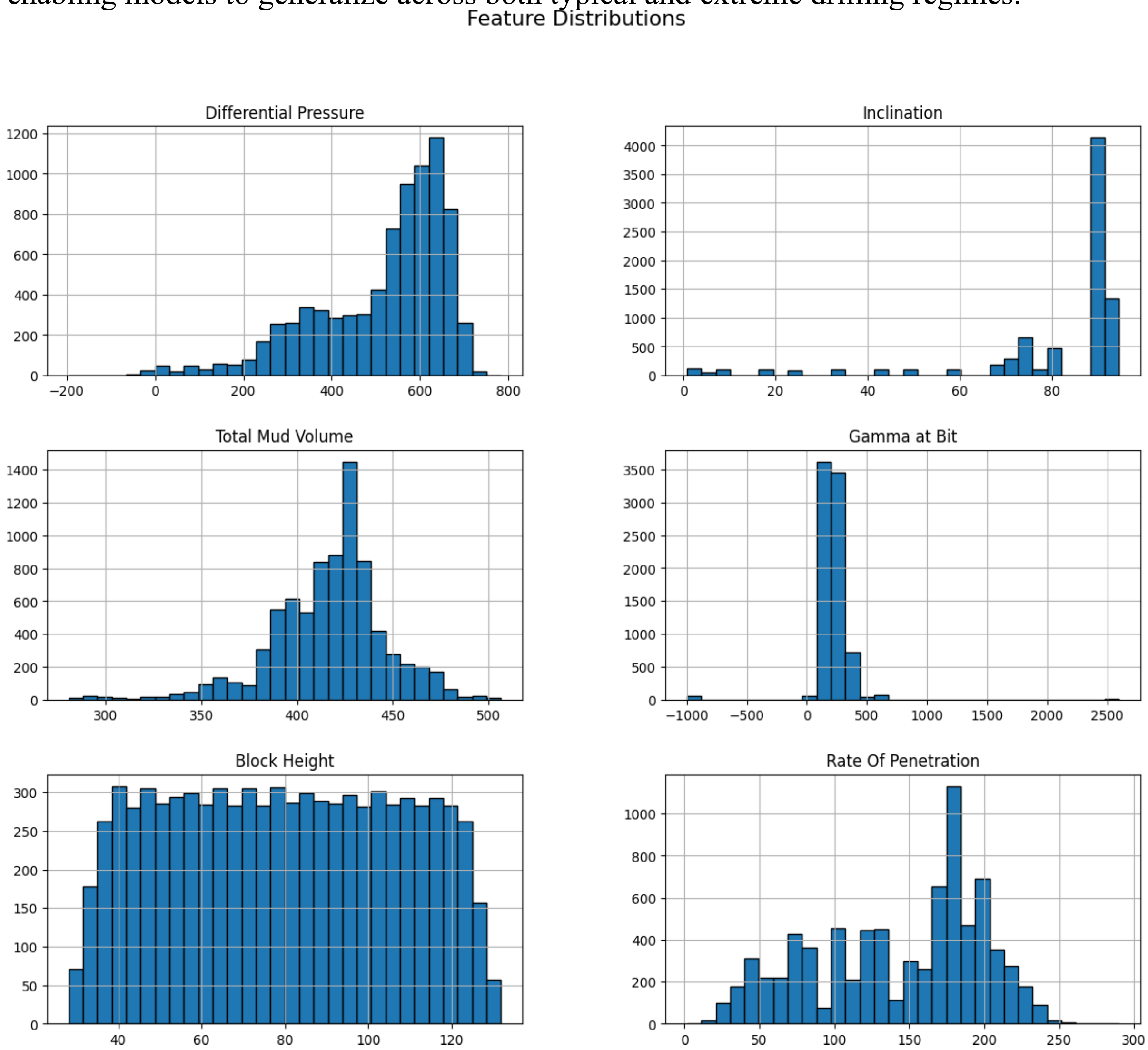

Figure 3. Histograms of key drilling features after preprocessing, Differential Pressure, Inclination, Total Mud Volume, Gamma at Bit, Block Height, and ROP, illustrating their distributions and variability for model training.

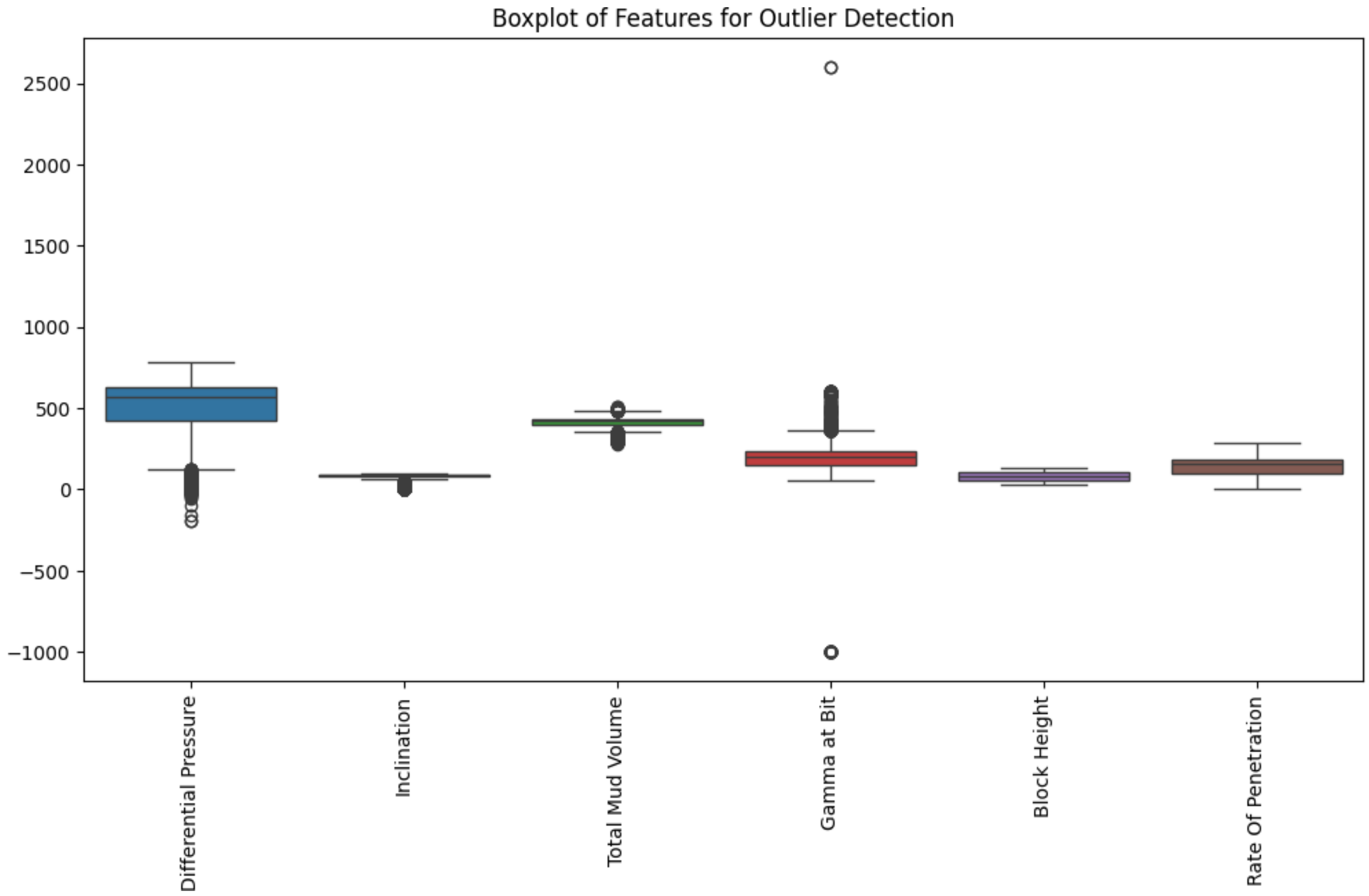

Figure 4. Boxplot of drilling features for outlier detection using the IQR method, with extreme values retained based on domain expert input.

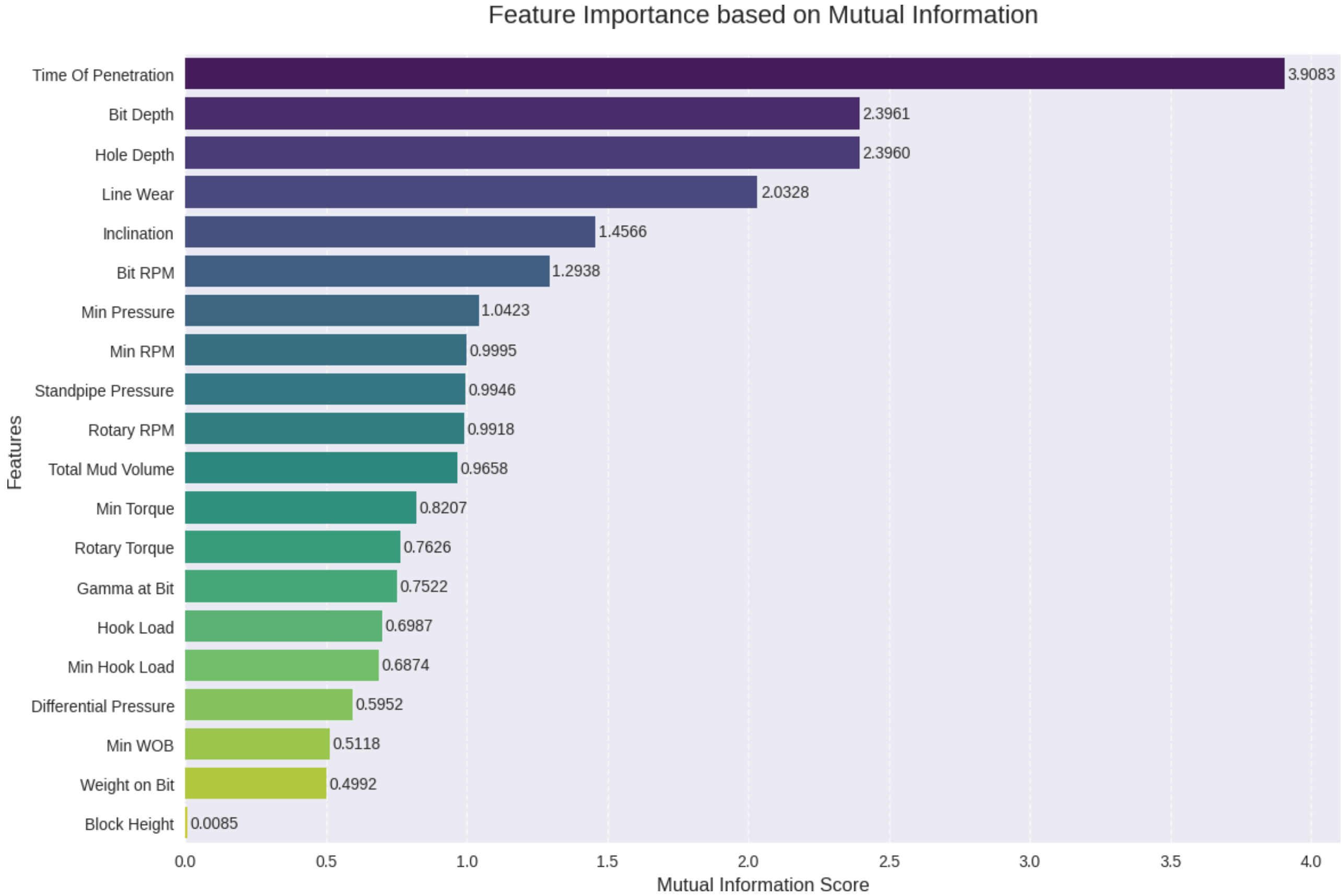

Figure 5. Feature importance ranking based on mutual information scores, highlighting key drilling parameters influencing model performance post.

Table 3. Data Preprocessing Pipeline Steps.

| Step No. | Step Description | Purpose |
|---|---|---|
| 1 | Missing Value Handling | Filled missing values using mean imputation to ensure data completeness and avoid training bias. |
| 2 | Feature Scaling | Applied StandardScaler normalization to standardize features for stable model convergence |
| 3 | Feature Selection | Retained relevant drilling parameters, excluded redundant or highly correlated features. |
| 4 | Target Variable Scaling | Standardized the target (ROP) to match feature scaling for better model fitting. |
| 5 | Train-Test Split | Divided the dataset into 80% training and 20% testing sets using random state = 42 for reproducibility. |
| 6 | Tensor Conversion | Converted all numpy arrays to PyTorch tensors for compatibility with deep learning Models. |
| 7 | DataLoader Creation | Batched the datasets (batch size = 64) and shuffled the training set to enhance the model's Generalization. |

There is no heavy feature engineering at first, to is conducive to the learning of the model. This choice was deliberate to save the model the end-to-end learning ability and not add predefined bias. But whenever they could, they also focused on levels and featurization of the product, such as:

a. Specific Energy Ratio = Torque * RPM / (WOB * ROP)

b. Hydraulic Horsepower = (GPM * PSI) / 1714 Where: GPM: Gallons Per Minute PSI: Pounds per square inch 1714

Preliminary analysis indicated that derived features did not outperform models trained on raw sensor data; therefore, direct sensor measurements were adopted as final inputs. Categorical variables, including Bit Type and BHA configuration, were transformed using one-hot encoding, producing binary representations suitable for neural network processing without implying ordinal relationships. Although feature dimensionality increased slightly, it remained manageable due to the limited number of categories. The ranking of influential drilling parameters from mutual information analysis is presented in Figure 5, highlighting features with the greatest impact on ROP prediction. All input and target variables were standardized using the formula $\mathbf{X}_{\text{scaled}}=(\mathbf{X}-\mu)/\sigma$, *where* $\mu$ and σ denote the mean and standard deviation, respectively [39]. Independent standardization of input features and ROP targets ensured zero mean and unit variance, reducing numerical instability and accelerating convergence during training. The fitted scaler objects were serialized using joblib to maintain consistent preprocessing for unseen data during deployment.

## 3.3 Deep Learning Integrated Framework for ROP Prediction

The proposed “Hybrid LSTM-Trans-Mixer-Att” DL framework extends customized LSTM networks to capture temporal dependencies, reduce vanishing gradient problems, and reveal hidden sequential patterns within drilling data. The architecture combines customized LSTM layers, improved Transformer encoders, Time-Series Mixer (TS-Mixer) blocks, and adaptive attention mechanisms to jointly model temporal continuity, overall context, static feature interactions, and dynamic feature significance. The customized LSTM module encodes multi-scale temporal patterns, effectively capturing both gradual operational trends and sudden transients. The enhanced Transformer encoder refines these representations with depth- and time-aware positional encodings, multi-head self-attention, and lightweight feed-forward layers, enabling efficient modeling of long-range dependencies across all time steps simultaneously. Meanwhile, the TS-Mixer projects categorical and continuous features into a high-dimensional latent space, applies MLP to boost cross-feature interactions, and outputs normalized static encodings.

The outputs of the Transformer and TS-Mixer are concatenated and processed through an adaptive attention mechanism, which calculates importance scores over temporal outputs to highlight key operational intervals. The resulting attention-weighted temporal vector, combined with the static encoding, is fed into a dense regressor to generate normalized ROP predictions. To ensure strong generalization, the framework uses dropout, batch normalization, MSE loss, and validation-based hyperparameter tuning. The customized LSTM is optimized with empirically selected parameters. Meanwhile, the Transformer is configured with four attention heads and a feed-forward dimension, and the TS-Mixer is tuned for efficient static feature modeling. Unlike traditional sequential LSTMs, the Transformer captures global dependencies across all time steps, while adaptive attention selectively emphasizes discriminative segments, especially those operational events that have a disproportionate impact on ROP.

### 3.3.1 Customized LSTM for Multi-Scale Temporal Feature Extraction

In this work, the LSTM block has been customized to align input sequences with drilling operational cycles, enabling domain-specific temporal representation and optimized for the target level of representation. A stacked configuration with two LSTM layers, each containing 64 hidden units, is used to capture multi-scale dependencies while maintaining a balance between representational capacity and computational efficiency (Figure 6). Additionally, regularization between layers mitigates overfitting on noisy drilling logs. The output sequence, representing aggregated temporal dynamics, is flattened and forwarded to the Transformer encoder for contextual refinement. In the attention-augmented hybrid model, the most advanced design incorporates a Transformer encoder to strengthen sequence modeling. The model first processes inputs $\boldsymbol{X} \in R^{N \times D}$ through a customized LSTM, whose outcomes are subsequently refined through an enhanced Transformer encoder, where multi-head self-attention and feed-forward layers capture global dependencies and complex feature interactions, thereby strengthening sequential drilling representation.

The architecture begins with the LSTM block, which processes the sequential input data step-by-step to capture temporal dependencies crucial for time-series tasks such as predicting drilling metrics. At each time step $t$, the LSTM cell receives three inputs: the current feature vector $X_t$, the previous hidden state $h_{t-1}$, and the previous cell state $c_{t-1}$. The cell operates through a system of gates that regulate memory update and information flow. Specifically, the input gate $i_t$ Determines how much new information from the current input should be added to the cell state, while the forget gate $f_t$ decides how much past information should be retained. The candidate cell state $g_t$ suggests new content to introduce, modulated by $i_t$. The current cell state $c_t$ is then updated as a weighted sum of the previous memory and the new candidate values, ensuring a balance between remembering and forgetting over time. Finally, the output gate $o_t$ controls the amount of information exposed to the next layers via the hidden state $h_t$. These operations are mathematically captured as:

$$i_t = \sigma(W_i \mathbf{X}_t + U_i h_{t-1} + b_i) \quad (1)$$

$$f_t = \sigma(W_f \mathbf{X}_t + U_f h_{t-1} + b_f) \quad (2)$$

$$o_t = \sigma(W_o \mathbf{X}_t + U_o h_{t-1} + b_o) \quad (3)$$

$$g_t = \tanh(W_g \mathbf{X}_t + U_g h_{t-1} + b_g) \quad (4)$$

$$c_t = f_t \odot c_{t-1} + i_t \odot g_t \quad (5)$$

$$h_t = o_t \odot \tanh(c_t) \quad (6)$$

Here, $\sigma$ is the sigmoid activation and $\odot$ denotes element-wise multiplication. The LSTM's internal gating mechanisms allow it to flexibly retain information over long sequences while adapting to new inputs, providing a rich temporal representation at each step.

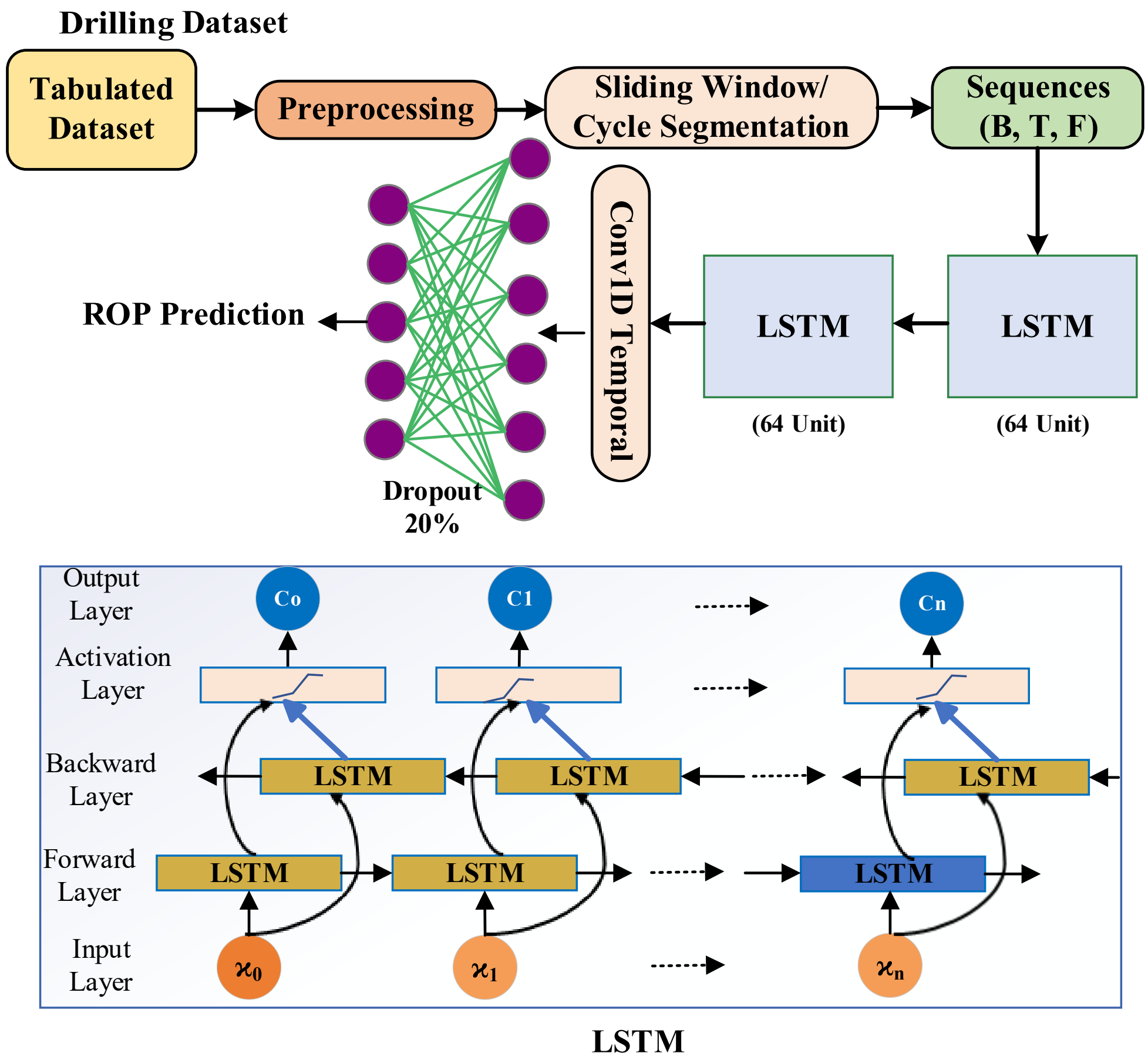

Figure 6. Data Pipeline from Tabular Dataset to LSTM Model Performance Assessment.

### 3.3.2 Enhanced Transformer for Global and Temporal Dependencies

The Transformer encoder has been enhanced for drilling-specific requirements by introducing customized positional encodings aligned with depth and time progression, as well as embedding layers. The embedding layers help in dimensionality reduction, and a shallow configuration is optimized for real-time deployment. The customized LSTM-derived outputs have been forwarded to the enhanced Transformer encoder, where global self-attention refines sequential dependencies and captures long-range relationships across all time steps (Figure 7). The LSTM output sequence $H_{\text{LSTM}}$ is first linearly transformed into a query $Q$, key $K$, and value $V$ matrices as seen in Eq. (7), where $W^Q$, $W^K$, and $W^V$ are learnable parameter matrices. The attention mechanism computes a weighted sum of values where the weights measure the compatibility between queries and keys, scaled by the dimension $d_k$ to maintain stable gradients:

$$Q = H_{\text{LSTM}} W^Q, K = H_{\text{LSTM}} W^K, V = H_{\text{LSTM}} W^V \qquad (7)$$

$$\text{Attention}(Q, K, V) = \text{softmax}\left(\frac{QK^\top}{\sqrt{d_k}}\right) V \qquad (8)$$

$$Y_{\text{ffn}} = \text{ReLU}(\mathbf{W}_1 Y_{\text{attn}} + b_1)\boldsymbol{W}_2 + b_2 \qquad (9)$$

$$Y_{\text{transformer}} = \text{LayerNorm}(Y_{\text{attn}} + Y_{\text{ffn}}) \quad (10)$$

Multiple attention heads operate in parallel, allowing the model to attend to various aspects of the sequence simultaneously. The concatenated output from these heads undergoes a feed-forward network comprising two fully connected layers with a ReLU non-linearity, enhancing expressiveness. Eq. 10 followed by residual connections and layer normalization to stabilize training. Through this block, the model refines the temporal representations by integrating global sequence information, focusing on the interactions between any distant time steps.

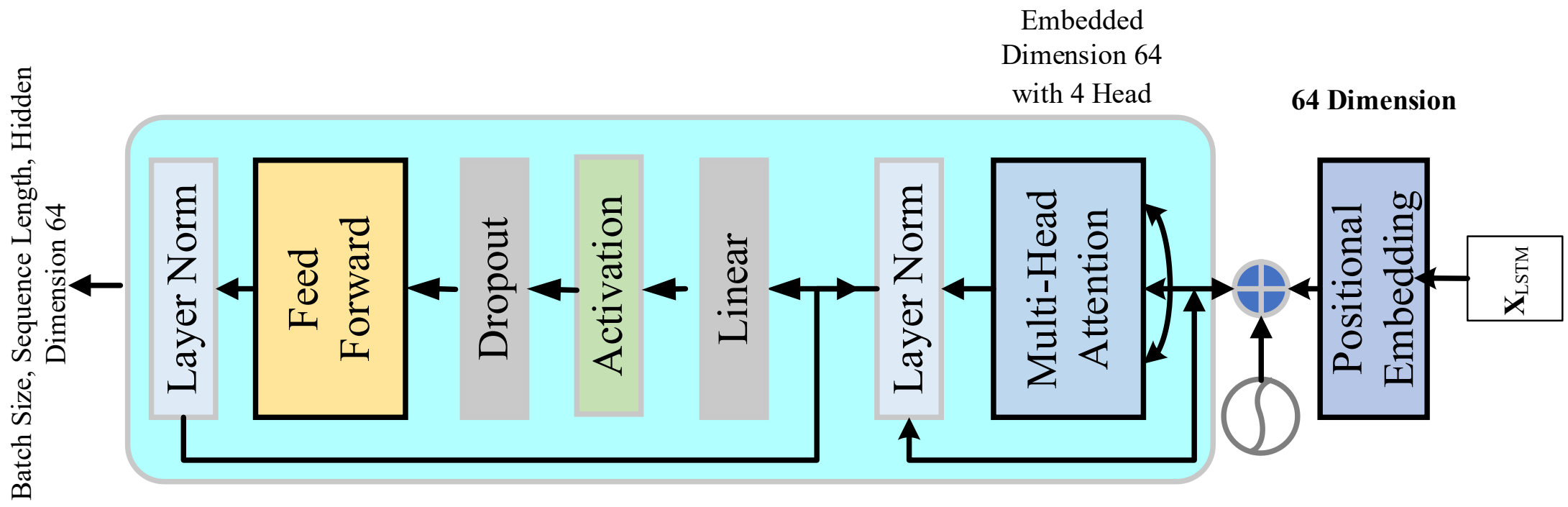

Figure 7: Enhanced Transformer encoder with drilling-specific positional encodings and LSTM embeddings for efficient long-range dependency modeling.

### 3.3.3 TS-Mixer Block Capture Static and Categorical Feature

The TS-Mixer is introduced for the first time in ROP prediction to capture nonlinear dependencies among static and categorical drilling attributes such as bit type, lithology indices, and mud properties. Unlike recurrent or attention-based architectures, it employs an alternating time- and feature-mixing MLP framework, where temporal and cross-feature interactions are modeled through fully connected transformations. This design is particularly effective for attributes that remain constant during drilling but exert significant contextual influence on dynamic operational behavior. By learning higher-order feature interactions often underutilized by LSTM and Transformer models, the TS-Mixer broadens the representational capacity and enhances generalization across diverse drilling environments. The architecture projects inputs into a 128-dimensional latent space, refined through stacked mixer layers with batch normalization, ReLU activation, residual connections, and dropout for stability and noise robustness. A temporal projection layer aligns input and forecast horizons, while the final projection generates normalized ROP estimates. By efficiently integrating static contextual information with sequential modules such as LSTM and Transformer layers, the TS-Mixer provides a lightweight yet expressive mechanism that improves prediction accuracy, mitigates overfitting, and enhances adaptability across drilling conditions. Mathematically, given the static feature vector $\mathbf{X}$, the output of the TS-Mixer's first layer is:

$$\mathbf{X}_{\text{mixer1}} = \text{ReLU}(\mathbf{W}_1\mathbf{X} + b_1) \quad (11)$$

$$\mathbf{X}_{\text{mixer2}} = \text{ReLU}(\mathbf{W}_2\mathbf{X}_{\text{mixer1}} + b_2) \quad (12)$$

where $W_1$ is a weight matrix of size $128 \times n$ (with $n$ being the number of static feature inputs), and $b_1$ is the associated bias. This intermediate representation is then passed to the second layer. Moreover, $W_2$ has dimensions $64 \times 128$, and $b_2$ is the bias for this layer. Each ReLU operation introduces nonlinearity, which helps the model learn complex inter-feature relationships. As a result, $\mathbf{X}_{\text{mixer2}}$ serves as a compact, high-level encoding of the static features suitable for fusion with learned temporal information in downstream layers.

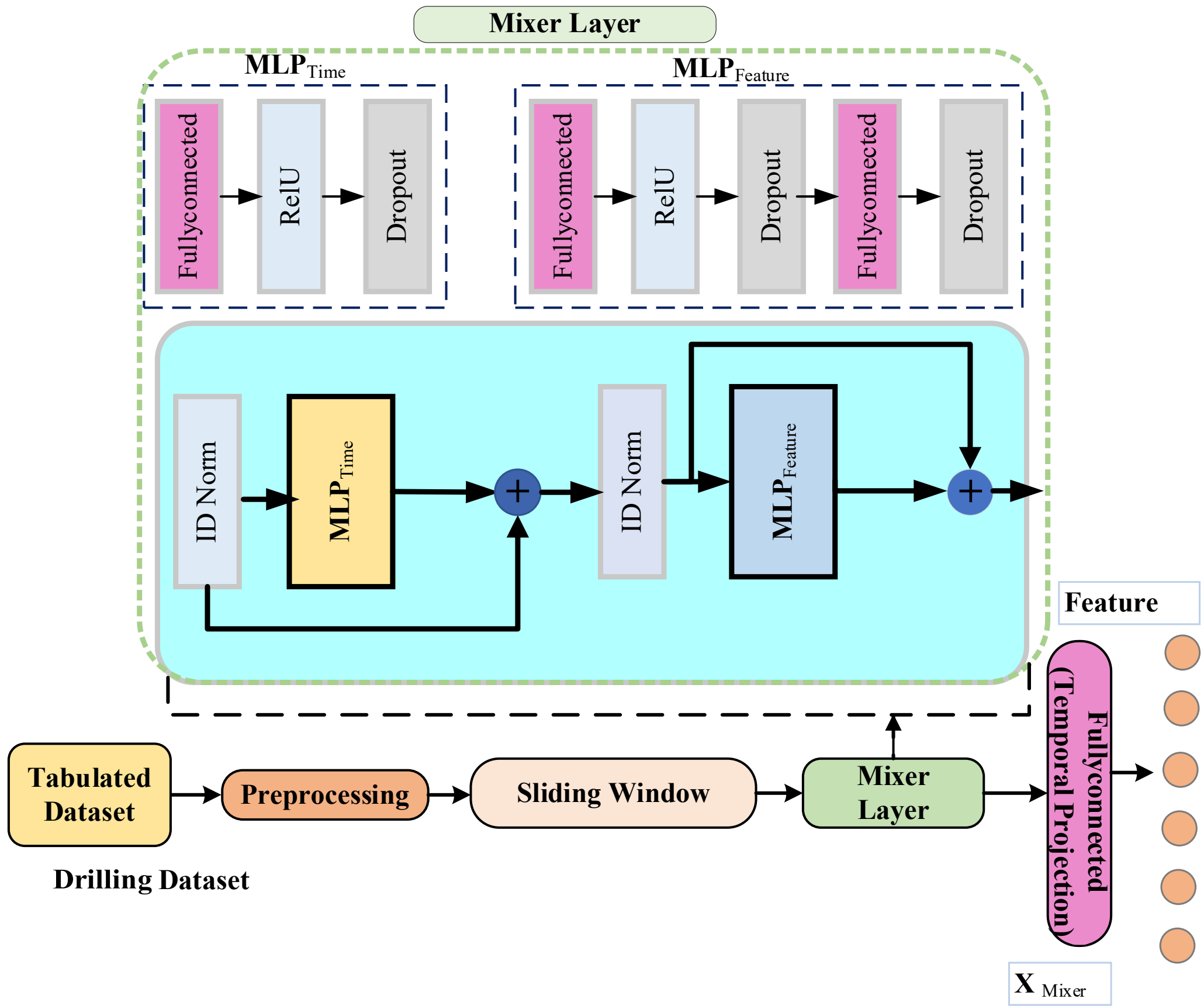

Figure 8. TS-Mixer Architecture for Predicting Rate of Penetration from Scaled Data.

### 3.3.4 Proposed Deep Boosted Feature with Adaptive Attention

The proposed "Hybrid LSTM-Trans-Mixer-Att" framework adopts a staged strategy for ROP prediction, progressing from baseline models to a unified hybrid architecture. The hybrid framework integrating customized LSTM, enhanced Transformer, and TS-Mixer demonstrated strong predictive capability. The attended temporal vector and global dependencies from $Y_{\text{Transformer}}$ are concatenated with the TS-Mixer output $\mathbf{X}_{\text{mixer2}}$, forming a unified representation that is processed by an adaptive attention mechanism with residual enhancement. The design is well-suited for drilling datasets, where abrupt operational shifts and heterogeneous feature distributions demand selective focus for accurate ROP prediction. The framework merges global context and boundary-aware information through a multilevel Transformer backbone, combining multiscale features across hierarchical stages and refining them with attention [40]. Sequential nonlinear activations generate pixel-level weights that reallocate feature relevance, enabling the model to capture both global dependencies and local variations.

The boosted feature representation is formulated in Eq. (13), where the attended temporal vector and global dependencies from $Y_{\text{Transformer}}$ are concatenated with the output of the TS-Mixer block $\mathbf{x}_{\text{mixer2}}$. This unified representation is then processed through an adaptive attention mechanism with residual enhancement, ensuring diverse yet coherent feature integration. Formally, the concatenated boosted features are expressed as $\mathbf{X}_{\text{boosted}}$ in Eq. (13). The refined activation $\mathrm{X}_{relu}$ is computed in (Eq. 15), followed by a nonlinear transformation to obtain pixel-level weights $\mathbf{W}_{fc}$(Eq. 14 & 16), where $\mathbf{W}_{\text{fc}}$ and $b_{\text{fc}}$ are the output layer's learnable parameters. The final attended representation is a weighted sum across time, aggregating temporal information according to learned task-relevant importance. The reweighted output $\mathbf{X}_{out}$ is given by (Eq. 14). Finally, temporal aggregation $e_t$ produces task-relevant embeddings (Eq. 17), with classification probabilities computed through Softmax (Eq. 18). This mechanism jointly models static and dynamic dependencies, improving robustness to abrupt lithological transitions while enhancing interpretability, accuracy, and real-time applicability in ROP prediction. Beyond predictive improvements, the framework incorporates preprocessing, regularization, and interpretability modules (SHAP, LIME, and attention visualization), ensuring transparent and transferable insights for ROP prediction and broader time-series forecasting applications.

$$\mathbf{X}_{\text{boosted}=}\text{b}(\ \mathbf{X}_{\text{Transformer}}||\ \mathbf{X}_{\text{Mixer}}) \tag{13}$$

$$\mathbf{X}_{out} = W_{fc}.\mathbf{X}_{boosted} \tag{14}$$

$$\mathbf{X}_{relu} = \sigma_1(\boldsymbol{W}_{\text{c}}\mathbf{X}_{Boosted} + b_{fc}) \tag{15}$$

$$\mathbf{W}_{fc} = \sigma_2(f(\mathbf{X}_{relu}) + b_f) \tag{16}$$

$$e_t = \sum_{\text{a}}^{\text{A}}\sum_{\text{b}}^{\text{B}}\text{v}_{\text{a}}\,\mathbf{X}_{out} \tag{17}$$

$$\sigma(\boldsymbol{d}) = \frac{\exp\,(e_t)}{\sum_{k=1}^{T}\exp\,(e_k)} \tag{18}$$

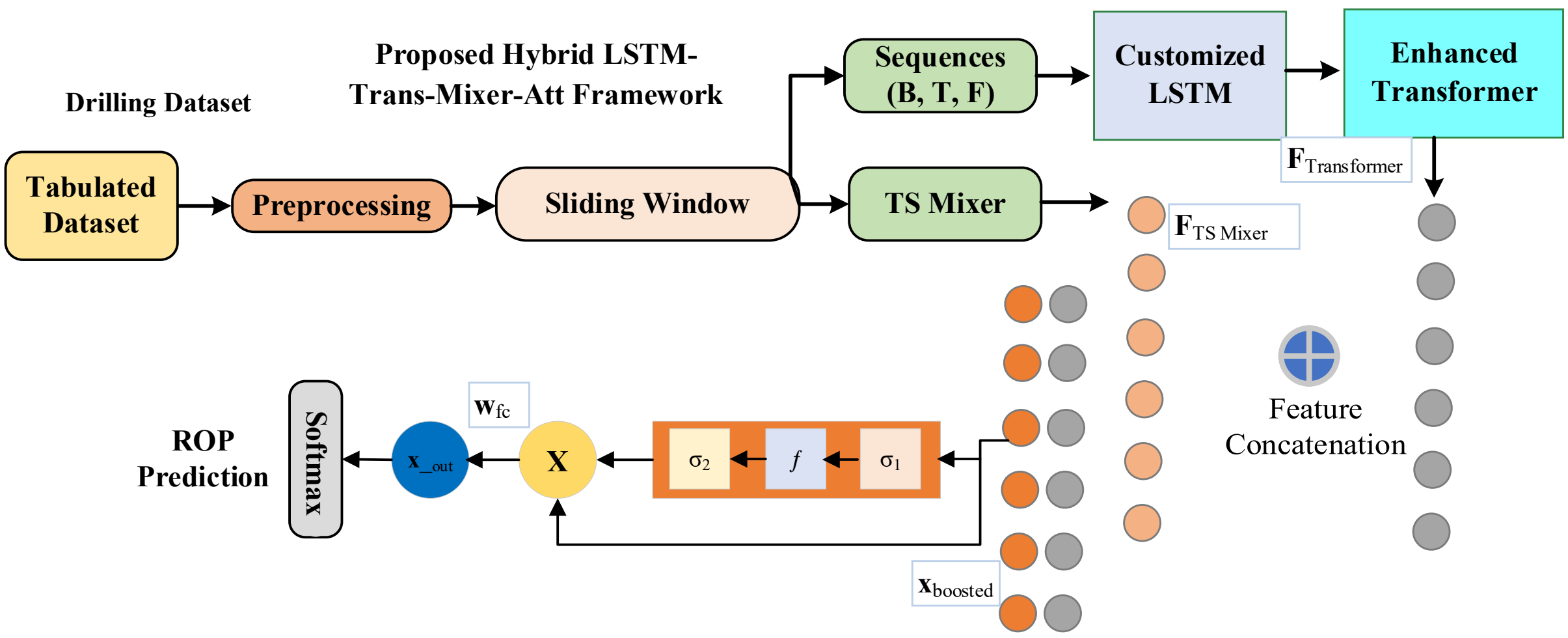

Figure 9. Multi-feature fusion and adaptive attention module in the proposed integrated framework.

### 3.3.5 Summary and Significance of the Proposed Framework

The proposed "Hybrid LSTM-Transformer-Mixer-Att" framework integrates a customized LSTM, enhanced Transformer, TS-Mixer, and adaptive attention within a hierarchical fusion pipeline optimized for the complex dynamics of drilling operations. A customized LSTM first captures multi-scale temporal dependencies aligned with drilling cycles. The enhanced Transformer, equipped with drilling-specific positional encodings and global self-attention, models long-range dependencies and contextual correlations. In parallel, the TS-Mixer, introduced for the first time in ROP prediction, efficiently captures nonlinear interactions among static and categorical features such as lithology, mud properties, and bit configuration.

The outputs of the Transformer and TS-Mixer are concatenated to form boosted feature representations, refined by an adaptive attention mechanism that applies dual-level prioritization to emphasize critical drilling intervals and salient attributes. The multilevel Transformer backbone fuses global context with boundary-aware information, while nonlinear activations generate adaptive weights for selective feature refinement, jointly capturing global dependencies and local variations.

Each module contributes a clearly defined functional role within the unified architecture:

- Customized LSTM ensures temporal continuity across drilling cycles.
- Enhanced Transformer models long-range global dependencies with contextual awareness.
- TS-Mixer captures nonlinear static–categorical feature interactions, strengthening cross-domain coherence.
- Adaptive Attention dynamically reweights salient drilling attributes, highlighting critical operational dynamics.

The integration of SHAP, LIME, and attention visualization enhances transparency and trustworthiness, while the modular design ensures scalability and adaptability across diverse time-series forecasting domains.

# 4 Experimental Setup

## 4.1 Implementation Setup

The DL models have been implemented in PyTorch with preprocessing handled through scikit-learn and pandas. Missing values were imputed, features standardized with StandardScaler, and inputs normalized using a combination of min–max scaling and z-score standardization. The dataset was divided into training and testing sets in an 80:20 ratio using train_test_split with a fixed random seed (42) to ensure reproducibility. Since ROP is a continuous target, stratification was not applied. Training data were shuffled at every epoch to minimize order bias.

Model training employed the AdamW optimizer (Loshchilov and Hutter, 2017) with an initial learning rate of 0.001, weight decay of $1 \times 10^{-5}$, a batch size of 64, and 100 epochs. Dropout (rate = 0.2) was applied for regularization, while PyTorch's default uniform parameter initialization was used. Mean Squared Error (MSE) served as the loss function, reflecting its suitability for regression and sensitivity to large deviations in ROP values. The architecture incorporated an adaptive attention mechanism, which improved convergence stability by producing smoother loss curves and mitigating overfitting. Early stopping was not applied to capture complete training dynamics, with potential overfitting monitored through validation loss trends.

Experiments were executed on an NVIDIA Tesla T4 GPU to enable efficient training and inference. Model persistence was ensured using PyTorch's torch. save for network parameters and joblib for preprocessing objects, supporting reproducible deployment. Random seeds were fixed across NumPy, scikit-learn, and PyTorch to control stochastic effects from initialization and shuffling. A summary of common hyperparameters is provided in Table 4: Optimizer = AdamW, Learning Rate = 0.001, Weight Decay = $1 \times 10^{-5}$, Batch Size = 64, Epochs = 100, and Dropout Rate = 0.2.

Table 4: Model-specific hyperparameters

| Model Name | LSTM Hidden Size | Number LSTM | Transformer Heads | Transformer FFN Size |
|---|---|---|---|---|
| Customized LSTM | 64 | 2 | - | - |
| TS-Mixer Model | - | - | - | - |
| Hybrid LSTM-TS-Mixer | 64 | 2 | - | - |
| Hybrid LSTM-TS-Mixer-Att | 64 | 2 | - | - |
| Proposed Hybrid Hybrid LSTM-Trans-Mixer-Att Framework (Final Model) | 64 | 2 | 4 | 128 |

Table 5: Summary of Model Architectures and Their Components

| Model Name | Components | Key Layers and Blocks | Output Dim. | Notable Features |
|---|---|---|---|---|
| Baseline LSTM Model | LSTM Network | LSTM (2 layers, hidden size 64), FC Layer | 1 | Sequence modeling, baseline performance |
| TS-Mixer Model | Feedforward Mixer | Dense(128), ReLU, Dense(64), ReLU, FC | 1 | Captures global dependencies, non-sequential |
| Hybrid LSTM + TS-Mixer Model | LSTM + TS-Mixer | LSTM (2 layers, 64 units) + TS-Mixer (Dense layers) + FC | 1 | Combines sequence features and global mixing |
| Hybrid LSTM + TS-Mixer + Attention Model | LSTM + TS-Mixer +Attention | LSTM (2 layers), Attention Layer, TS-Mixer, FC | 1 | Focused learning through dynamic feature weighting |
| Baseline LSTM Model | LSTM Network | LSTM (2 layers, hidden size 64), FC | 1 | Sequential + relational + global feature fusion |
| Proposed Hybrid LSTM-Trans-Mixer-Att Framework | LSTM + Transformer Encoder + TS-Mixer+ Attention | LSTM (2 layers), Transformer Encoder (4 heads), Attention, TS-Mixer, FC | 1 | Sequence modeling, baseline performance |

### 4.2 Evaluation Metrics

The performance of the regression models has been quantitatively assessed using multiple statistical metrics to ensure a comprehensive evaluation [41]. Standard regression metrics, including the coefficient of determination ($R^2$), Mean Absolute Error (MAE), Root Mean Square Error (RMSE), and Mean Absolute Percentage Error (MAPE), were employed to rigorously evaluate model performance. All evaluations were performed on a held-out test set comprising 20% of the dataset. To maintain interpretability, both predicted and ground-truth Rate of Penetration (ROP) values were inverse-transformed to their original physical units using the fitted scaling parameters.

Reliance on a single performance index often obscures critical aspects of predictive reliability; therefore, a comprehensive set of metrics has been adopted to jointly capture absolute error magnitude, relative deviation, and explained variance.

1. The $R^2$ score of a model shows the proportion of the variance in the dependent variable that is predictable from the independent variables. Closer to 1 indicates better prediction performance.

$$R^2 = 1 - \frac{\sum_{i=1}^{n}(y_i - \hat{y}_i)^2}{\sum_{i=1}^{n}(y_i - \bar{y})^2}$$

2. MAE represents the average absolute difference between predicted and real ROP values. The smaller the MAE values, the greater the accuracy.

$$\text{MAE} = \frac{1}{n}\sum_{i=1}^{n}|y_i - \hat{y}_i|$$

3. This is the square root of the average of the squared differences between predicted and actual values and penalizes large errors more heavily than the MAE.

$$\text{RMSE} = \sqrt{\frac{1}{n}\sum_{i=1}^{n}(y_i - \hat{y}_i)^2}$$

4. MAPE measures the average forecast error as a percentage of the actual values, and it provides a normalized measure of accuracy.

$$\text{MAPE} = \frac{100\%}{n} \sum_{i=1}^{n} \left| \frac{y_i - \hat{y}_i}{y_i} \right|$$

# 5 Results

The proposed framework demonstrates state-of-the-art predictive accuracy, establishing a robust benchmark for next-generation ROP prediction architectures. By sequentially integrating customized LSTM units, enhanced Transformer encoders, concatenated TS-Mixer features, and adaptive attention mechanisms, it achieves superior feature representation and learning efficiency. Predictive capability is rigorously validated using $R^2$, MAE, RMSE, and MAPE metrics computed on inverse-transformed ROP values, ensuring evaluation fidelity.

The architecture emphasizes multi-general learning and resilient feature abstraction, offering a lightweight yet highly accurate model adaptable to diverse geological formations. Comparative experiments on real-world drilling datasets confirm consistent outperformance over baseline and hybrid models, including LSTM, TS-Mixer, LSTM–TS-Mixer, and LSTM–TS-Mixer–Att configurations (Figure 10).

Empirical analysis reveals a direct correlation between architectural complexity and predictive gain. Integrating customized LSTM layers with Transformer encoders, concatenated TS-Mixer blocks, and attention mechanisms significantly enhances accuracy, robustness, and generalization. As shown in Figure 10, each added component incrementally improves stability and predictive consistency, affirming the proposed hybrid framework's scalability and efficacy for intelligent drilling optimization.

## 5.1 The Proposed Hybrid DL Framework

The proposed hybrid DL framework, integrating customized LSTM layers, enhanced Transformer encoders concatenated with TS-Mixer blocks, and adaptive attention, achieved the highest predictive accuracy among all models. After 100 epochs, it attained an $R^2$ of 0.9991, MAE of 1.2531, RMSE of 1.6573, and MAPE of 1.1572%, indicating near-perfect alignment between predicted and ground-truth ROP values. The architecture demonstrated excellent generalization with no observable overfitting, confirming the effectiveness of its multi-component design. Each component contributed distinct advantages: Transformer encoders strengthened long-range dependency modeling, TS-Mixer blocks captured nonlinear static–sequential interactions, and attention adaptively weighted salient features, enhancing contextual representation. Together, these modules complemented the LSTM's temporal learning, forming a unified model capable of representing both local dynamics and global dependencies.

Model interpretability was examined using LIME and SHAP analyses. LIME consistently identified Time of Penetration as a strong negative contributor (–1.91 to –0.41), while Hole Depth, Hook Load, and Torque exhibited dominant positive effects (>0.8) (Figure 11). Negative influences were observed for Minimum RPM and Differential Pressure (–0.65 to –0.59), aligning closely with actual predictions (e.g., Ground Truth = 49.69, Prediction = 48.86). Global SHAP analysis corroborated these findings, highlighting Hole Depth and Hook Load as the most influential predictors and capturing nonlinear interactions among Torque, RPM, and Pressure. Moreover, Performance validation through density plots and residual error analysis (Figure 13) showed substantial overlap between predicted and actual ROP distributions, with absolute errors between 0.93 and 4.58. Despite minor deviations in extreme ROP regimes, predictive fidelity remained state-of-the-art. The observed bimodal distribution further indicated the model's capability to capture dominant operational states and transitional drilling phases, suggesting potential for targeted fine-tuning in high-range scenarios.

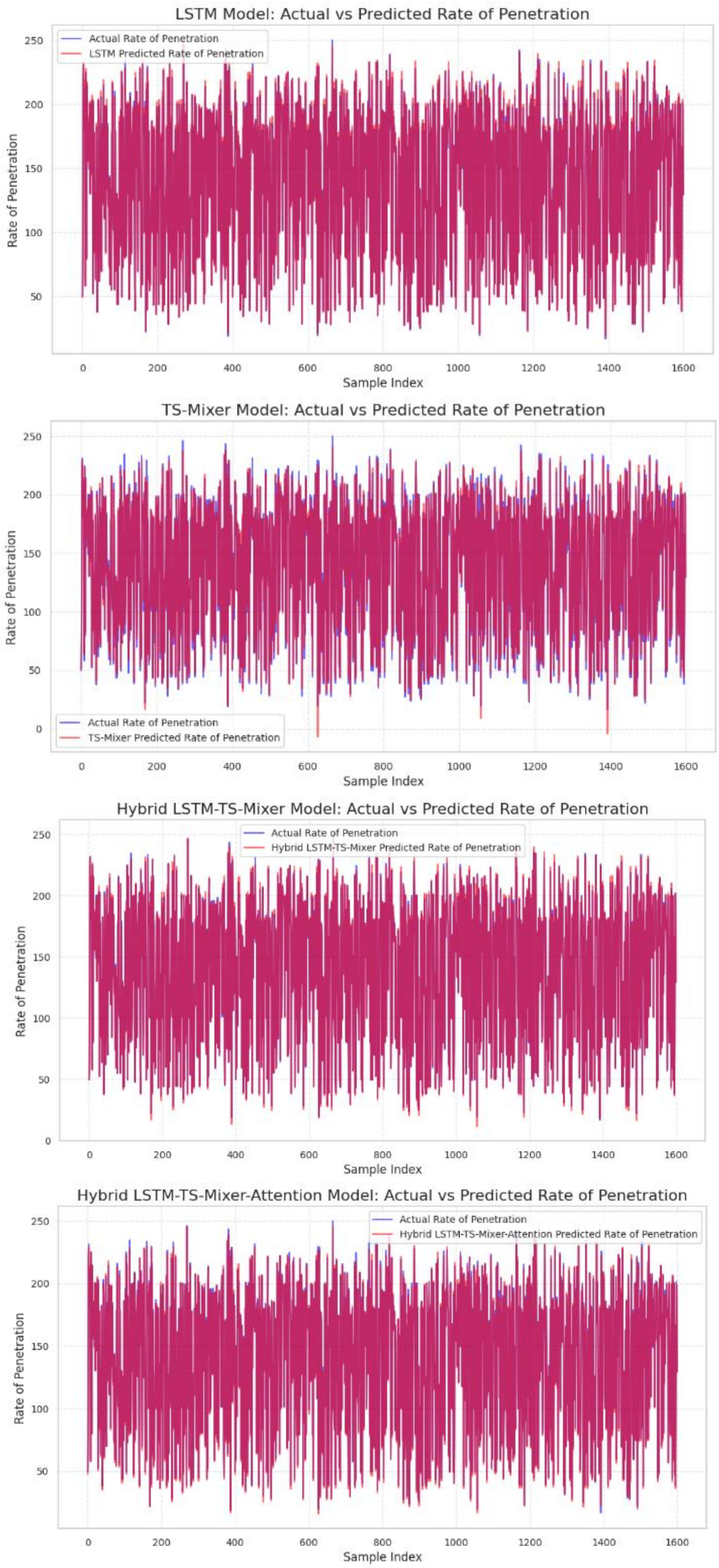

Figure 10: Actual vs. predicted ROP for Customized LSTM, TS-Mixer, and Hybrid LSTM–TS-Mixer–Att, illustrating progressive enhancement in learning nonlinear and temporal patterns.

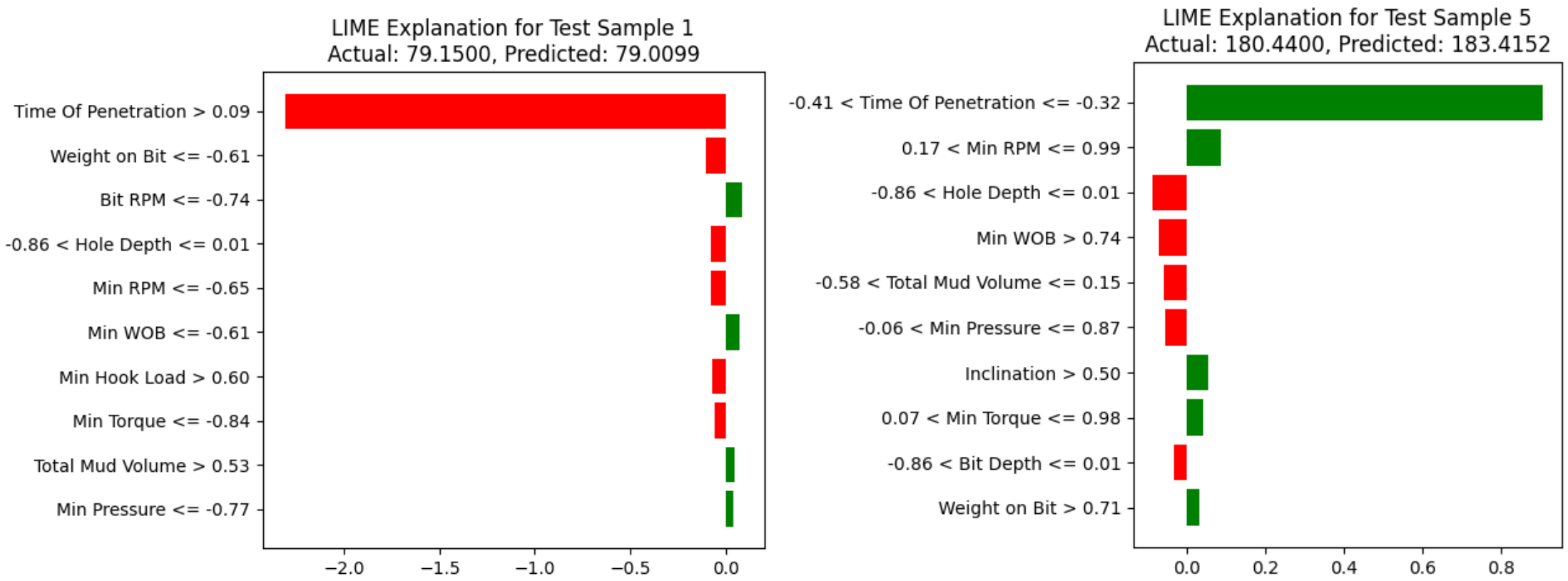

Figure 11. LIME Explanation of Feature Impact on Predicted ROP for Test Samples 1 and 5.

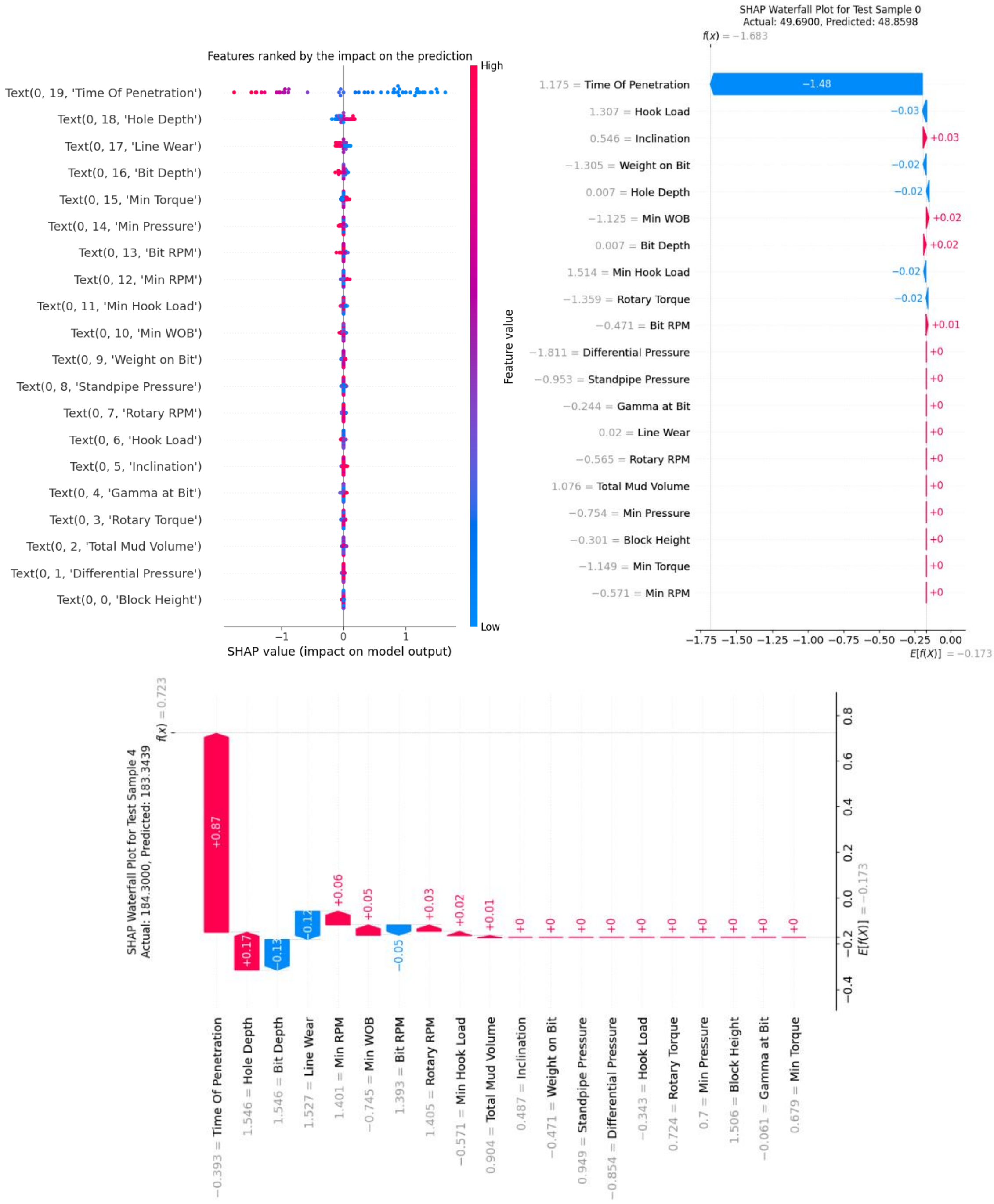

Figure 12. SHAP Feature Importance Ranking Based on Impact on Model Prediction.

Averaged SHAP values (Table 7, Figures 12) confirmed Time of Penetration as the strongest negative driver (–0.43), with minor contributions from Inclination (+0.03) and Torque (+0.01), and negligible effects from secondary parameters. These interpretability insights affirm that the framework not only achieves exceptional predictive precision but also preserves transparency through quantifiable feature attribution. Its capability to identify operational thresholds, model dynamic dependencies, and provide interpretable decision support establishes it as a robust, domain-agnostic tool for real-time drilling optimization and complex time-series forecasting.

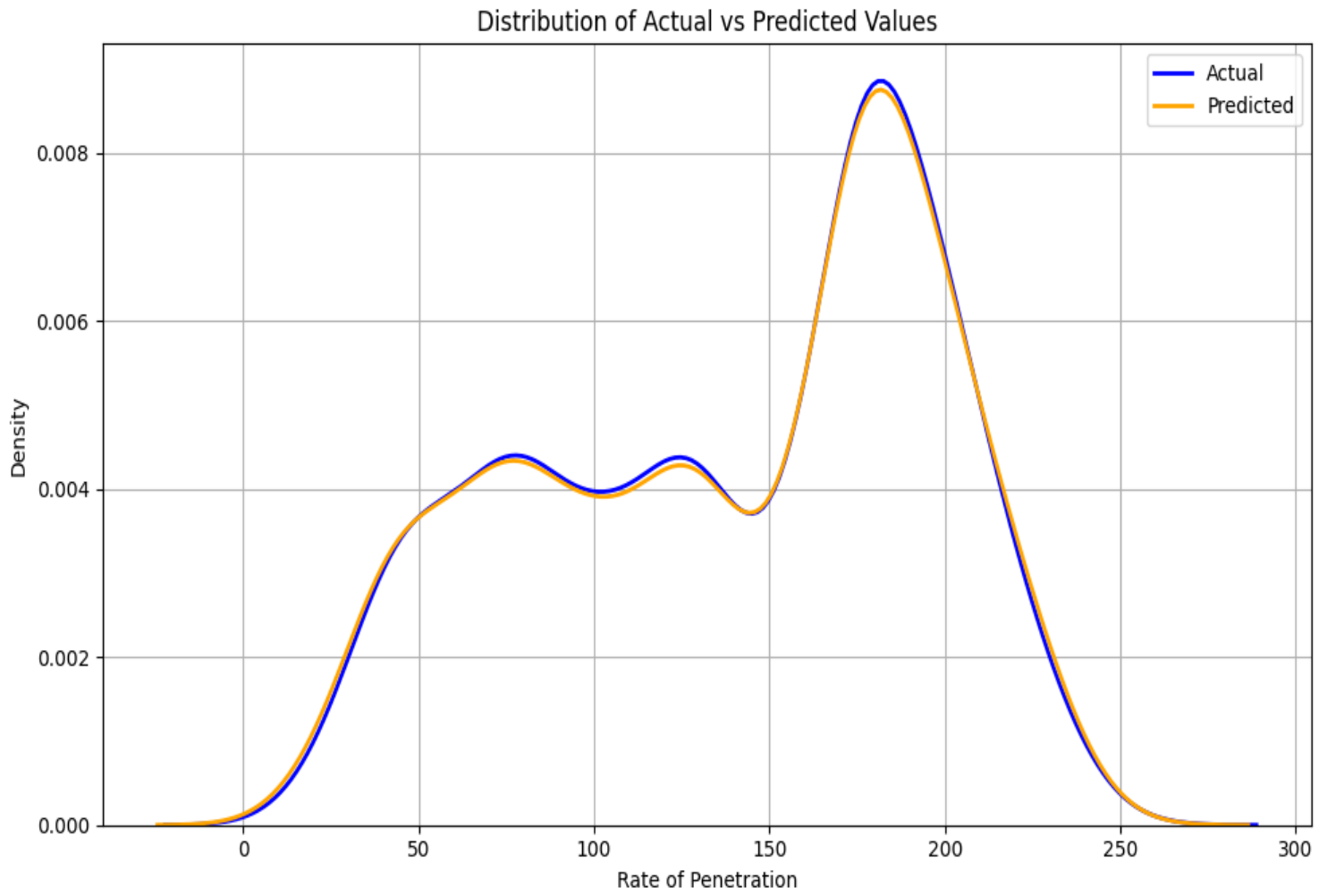

Figure 13. Density Distribution curve of Actual vs. Predicted ROP Values.

Table 6: Actual vs. Predicted Values and Absolute Errors for Test Data

| Samples | Actual Value | Predicted Value | Absolute Error |
|---|---|---|---|
| 0 | 49.6900 | 46.9230 | 2.7670 |
| 1 | 79.1500 | 78.2176 | 0.9324 |
| 2 | 218.6300 | 222.9723 | 4.3423 |
| 3 | 231.3800 | 230.4430 | 0.9370 |
| 4 | 184.3000 | 179.7132 | 4.5868 |
| 5 | 180.4400 | 182.0244 | 1.5844 |
| 6 | 211.3700 | 215.3411 | 3.9711 |
| 7 | 75.9000 | 74.7389 | 1.1611 |
| 8 | 58.1800 | 57.1802 | 0.9998 |
| 9 | 224.7400 | 226.6308 | 1.8908 |

Table 7: Feature Importance and Common Condition Thresholds with Corresponding Occurrences

| **Feature** | **Avg.Weight** | **Common Condition** | **Occurrences** |
|---|---|---|---|
| Time of Penetration | -1.09 | > 0.09 (Samples 0, 1, 7, 8) or ≤ -0.41 (Samples 2–6, 9) | 10 |
| Min WOB | -0.11 | ≤ -0.61 (Samples 0, 1) or > 0.74 (Samples 5, 7, 9) | 6 |
| Weight on Bit | -0.05 | ≤ -0.61 (Samples 0, 1) or > 0.71 (Samples 5, 7, 9) | 6 |
| Min Torque | -0.12 | ≤ -0.84 (Samples 0, 1) or > 0.98 (Samples 3, 6, 9) | 8 |
| Rotary Torque | 0.08 | ≤ -0.67 (Sample 0) or > 0.82 (Samples 3, 6) | 3 |
| Hole Depth | 0.01 | 0.01 < x ≤ 0.88 (Sample 0) or > 0.88 (Samples 3, 4, 6, 9) | 8 |
| Differential Pressure | -0.57 | ≤ -0.59 (Samples 0, 1, 4, 8) or ≤ -0.33 (Sample 6) | 5 |
| Min Hook Load | -0.06 | ≤ -0.83 (Samples 3, 6, 9) or > 0.60 (Samples 0, 1) | 7 |
| Line Wear | -0.06 | ≤ 0.02 (Sample 0) or > 0.86 (Samples 3, 6, 9) | 6 |
| Block Height | -0.18 | ≤ 0.00 (Samples 0, 3) or > 0.86 (Samples 2, 4) | 4 |
| Bit RPM | -0.09 | ≤ -0.74 (Samples 1, 8) or > 0.97 (Samples 3, 6) | 6 |
| Min RPM | -0.27 | ≤ -0.65 (Samples 1, 7, 8) or ≤ 0.99 (Samples 2–6, 9) | 9 |
| Inclination | 0.25 | > 0.50 (Samples 2, 4, 5) or ≤ 0.22 (Sample 7) | 4 |
| Total Mud Volume | 0.27 | > 0.53 (Sample 1) or ≤ 0.15 (Sample 5) | 2 |
| Min Pressure | -0.29 | ≤ -0.77 (Samples 1, 7) or > 0.87 (Sample 9) | 4 |
| Hook Load | -0.43 | ≤ -0.76 (Samples 3, 6, 7, 9) or ≤ -0.03 (Sample 7) | 5 |
| Rotary RPM | 0.29 | > 0.99 (Samples 4, 6) or ≤ 0.99 (Sample 2) | 3 |
| Bit Depth | 0.26 | > 0.88 (Samples 3, 9) or ≤ 0.01 (Samples 1, 5) | 5 |
| Standpipe Pressure | 0.88 | > 0.88 (Sample 6) | 1 |

## 5.2 Behavior of Training and Validation Loss Curves

All model configurations exhibited smoothly declining training and validation loss trajectories, with advanced architectures achieving consistently lower terminal losses. The baseline customized LSTM showed mild overfitting, evidenced by a slight divergence between training and validation losses during the final epochs (Figure 14). Conversely, the TS-Mixer and hybrid variants maintained tightly aligned loss curves, reflecting enhanced generalization and mitigated overfitting. The proposed Advanced Hybrid architecture, combining customized LSTM, enhanced Transformer encoder, TS-Mixer, and Attention, demonstrated the most stable convergence, characterized by a persistently minimal loss gap across epochs. This convergence stability, corroborated by low inter-run variance, confirms the model's reliability, efficient optimization, and reproducible learning dynamics.

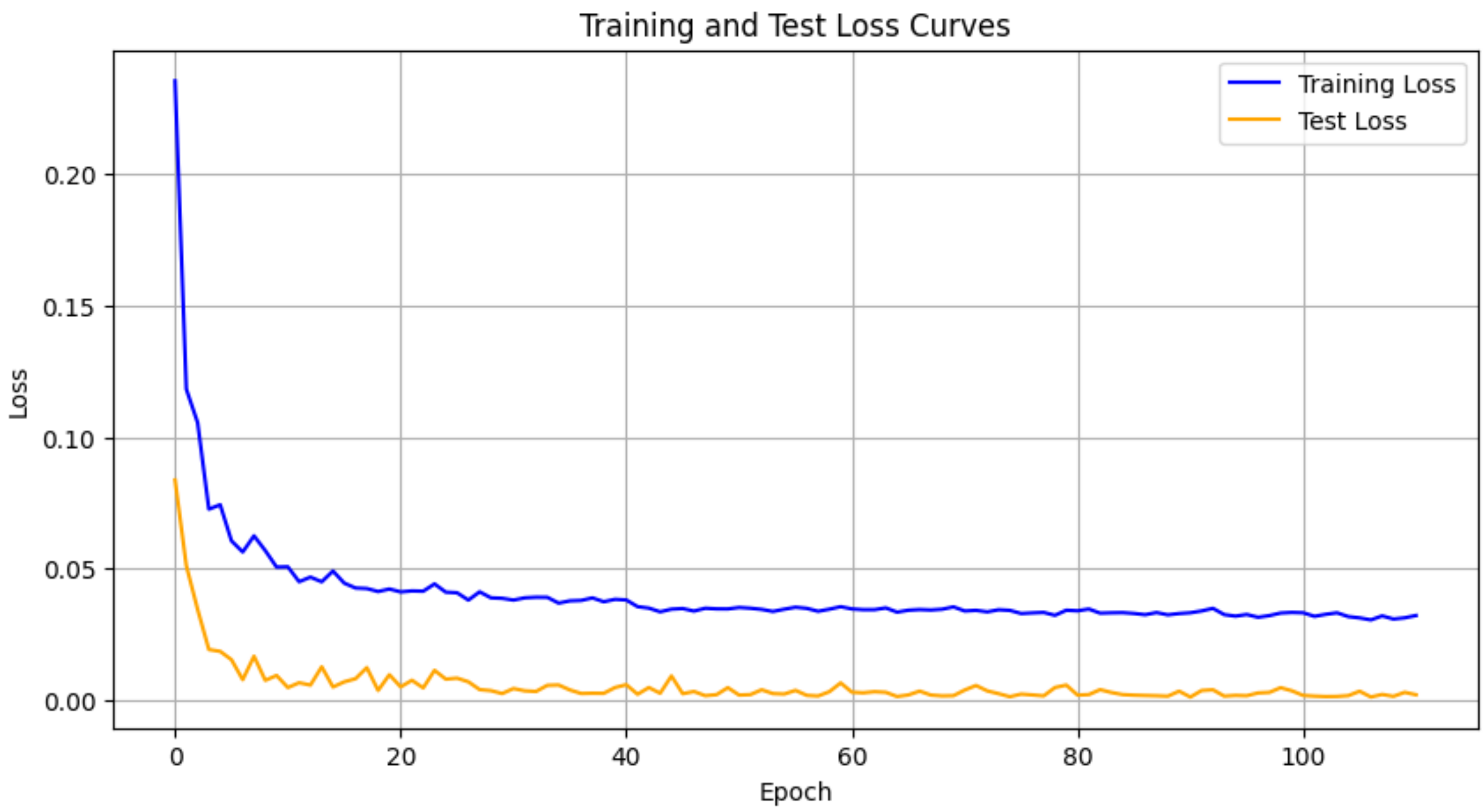

Figure 14. Training and Test Loss Curves Over 100 Epochs.

## 5.3 Ablation Study

A comprehensive ablation study was conducted to quantify the contribution of each architectural component, Customized LSTM, Enhanced Transformer encoder, TS-Mixer, and adaptive Attention, toward improving ROP prediction accuracy, stability, and generalization. The analysis progresses from individual models to the fully integrated hybrid framework, as detailed in Table 8. The customized LSTM achieved an $R^2$ of 0.9881, demonstrating strong temporal modeling capability across drilling sequences. Nonetheless, during abrupt ROP fluctuations, minor over- and underestimations emerged, indicating limited responsiveness to high-frequency transients and nonlinear dependencies. Although effective for sequential continuity, its shallow structure constrained generalization under complex drilling dynamics, motivating the integration of a Transformer encoder for global context refinement. The Enhanced Transformer, equipped with drilling-specific positional encodings and real-time optimization, refined the temporal features generated by the LSTM. This module captured long-range dependencies and enhanced global contextual awareness, mitigating the localized sensitivity inherent in recurrent architectures.

The TS-Mixer, implemented as a deep MLP for static and cross-feature interaction learning,

achieved an $R^2$ of 0.9845 with a MAPE of 4.833%. While slightly lower in accuracy than the LSTM, it demonstrated superior robustness and consistency, validating its role in modeling nonlinear dependencies among static and categorical drilling parameters such as lithological indices and mud properties. Subsequently, the fusion of Enhanced Transformer and TS-Mixer features, concatenated after LSTM processing, significantly boosted predictive reliability. This hybrid configuration achieved an $R^2$ of 0.9860, MAE of 2.257, RMSE of 2.778, and MAPE of 1.953%, reflecting balanced adaptability and improved generalization across heterogeneous drilling conditions. The close alignment between predicted and actual ROP curves confirmed effective temporal-static feature fusion.

Incorporating the adaptive Attention mechanism further enhanced interpretability and feature discrimination. The final LSTM-Transformer-Mixer-Att configurations, Customized LSTM with Enhanced Transformer ‖ TS-Mixer + Attention, achieved an $R^2$ of 0.9991, MAE of 1.2531, RMSE of 1.6573, and MAPE of 1.1572%. Attention selectively emphasized salient temporal and contextual dependencies while suppressing redundant information, ensuring resilience in scale-sensitive and noisy operational intervals. Overall, the ablation analysis revealed a consistent upward trajectory in performance with each added component. The synergy among sequential memory modeling, global context learning, static feature interaction, and adaptive attention weighting established the proposed hybrid framework as a technically robust and high-fidelity benchmark for intelligent ROP prediction systems.

Table 8: Comparative Performance Metrics of Different Models for ROP Prediction

| Model | $R^2$ Score | MAE | RMSE | MAPE | Flops | Inf.T (ms) | T/Ep (M) |
|---|---|---|---|---|---|---|---|
| LSTM Baseline | 0.9881 | 2.320 | 3.506 | 1.950% | ~3.1 | ~1.4 | ~2.3 |
| TS-Mixer | 0.9845 | 2.732 | 4.059 | 2.833% | ~2.6 | ~1.2 | ~2.0 |
| Hybrid LSTM + TS-Mixer | 0.9860 | 2.257 | 2.778 | 1.953% | ~4.1 | ~2.7 | ~4.3 |
| Hybrid LSTM + Transformer + Attention | 0.9907 | 1.830 | 2.150 | 1.525% | ~5.6 | ~3.3 | ~5.0 |
| Proposed DL Hybrid LSTM-Trans-Mixer-Att | 0.9991 | 1.253 | 1.657 | 1.157% | ~10.1 | ~4.7 | ~6.3 |

## 5.4 Comparative Analysis

The proposed DL hybrid ‘LSTM-Trans-Mixer-Att’ framework consistently outperformed all other models, achieving an $R^2$ of 0.9991, which explains nearly the entire variance in ROP predictions and clearly surpasses all baseline and hybrid counterparts (Table 8, Figure 15). In comparison, the baseline LSTM achieved an $R^2$ of 0.9881, the TS-Mixer reached 0.9845, the Hybrid LSTM + TS-Mixer obtained 0.9860, and the Hybrid customized LSTM + Transformer + Attention attained 0.9907, confirming progressive improvements from successive architectural integration.

In terms of error metrics, the proposed framework recorded the lowest MAE (1.253) and RMSE (1.657), demonstrating substantial reductions in prediction deviations compared to the Hybrid LSTM + TS-Mixer (MAE = 2.257, RMSE = 2.778) and standalone TS-Mixer (MAE = 5.732, RMSE = 7.059). Moreover, it achieved an MAPE of 1.157%, indicating an exceptionally high percentage-based precision and outperforming all preceding configurations, with only minor deviation from the Hybrid LSTM + Transformer + Attention (1.525%)

These empirical findings validate that the integration of enhanced Transformer encoders and adaptive attention mechanisms with customized LSTM and TS-Mixer blocks effectively captures both long-term temporal dependencies and nonlinear cross-feature interactions. Consequently, the framework demonstrates superior accuracy, stability, and generalization across heterogeneous drilling environments. The marginally elevated MAPE value reflects minimal sensitivity to percentage-based variations, suggesting that further hyperparameter optimization or geological domain-specific calibration could yield near-optimal performance.

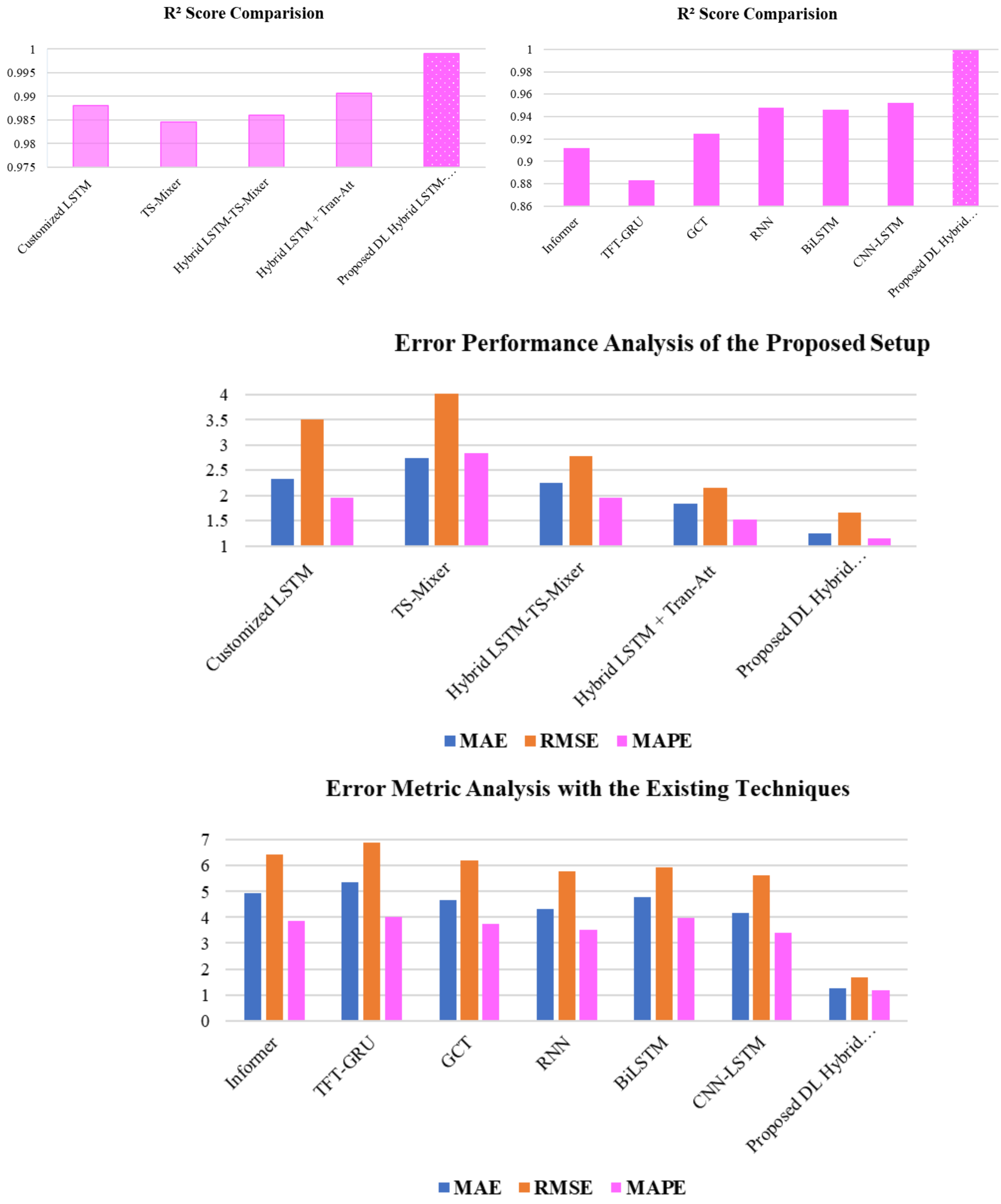

Figure 15. Model Performance Comparison Across Metrics with Error Bars.

## 5.5 Performance Comparison of Benchmark Models

A comprehensive comparative evaluation has been conducted between the proposed Advanced Hybrid framework and several established benchmark architectures for forecasting the S&P 500 Index, a canonical and challenging proxy used to examine model robustness and generalizability in financial time-series prediction (Figure 15, Table 9). Conventional DL baselines, including Informer [34], Temporal Fusion Transformer with GRU backbone (TFT-GRU) [42], Gated Convolutional Transformer (GCT) [43], RNN, BiLSTM, and CNN-LSTM [44], achieved moderate predictive performance, with $R^2$ values ranging from 0.883 to 0.952 and RMSE values between 5.63 and 6.88. Among these, CNN-LSTM ($R^2 = 0.952$, RMSE = 5.63) and RNN ($R^2 = 0.948$, RMSE = 5.79) achieved comparatively higher accuracy, reflecting their enhanced capability for temporal dependency modeling. Conversely, TFT-GRU ($R^2$ = 0.883, RMSE = 6.88) and Informer ($R^2$ = 0.912, RMSE = 6.43) exhibited larger prediction errors, highlighting their limited adaptability to nonlinear temporal fluctuations inherent in the drilling sequence.

In contrast, the proposed Advanced Hybrid framework (customized LSTM + Enhanced Transformer ||TS-Mixer) with Adaptive Attention achieved a significantly higher $R^2$ of 0.9991, along with notably reduced MAE (1.2531), RMSE (1.6573), and MAPE (1.1572%). These results confirm its superior ability to capture intricate short- and long-range temporal dependencies, adaptively emphasize critical operational intervals through dual-stage attention, and effectively mitigate gradient degradation issues that often limit conventional recurrent or attention-based models.

Figure 15 and Table 9 illustrate this marked improvement, where the predicted ROP trajectories from the proposed model align almost perfectly with the ground truth, contrasting sharply with the lagging and fluctuating responses of benchmark methods. The empirical findings substantiate that the proposed Hybrid framework delivers state-of-the-art precision and generalization across highly nonlinear and dynamic datasets. Its methodological synergy and adaptive fusion design position it as a robust, domain-agnostic solution for both industrial time-series prediction and broader sequential modeling applications.

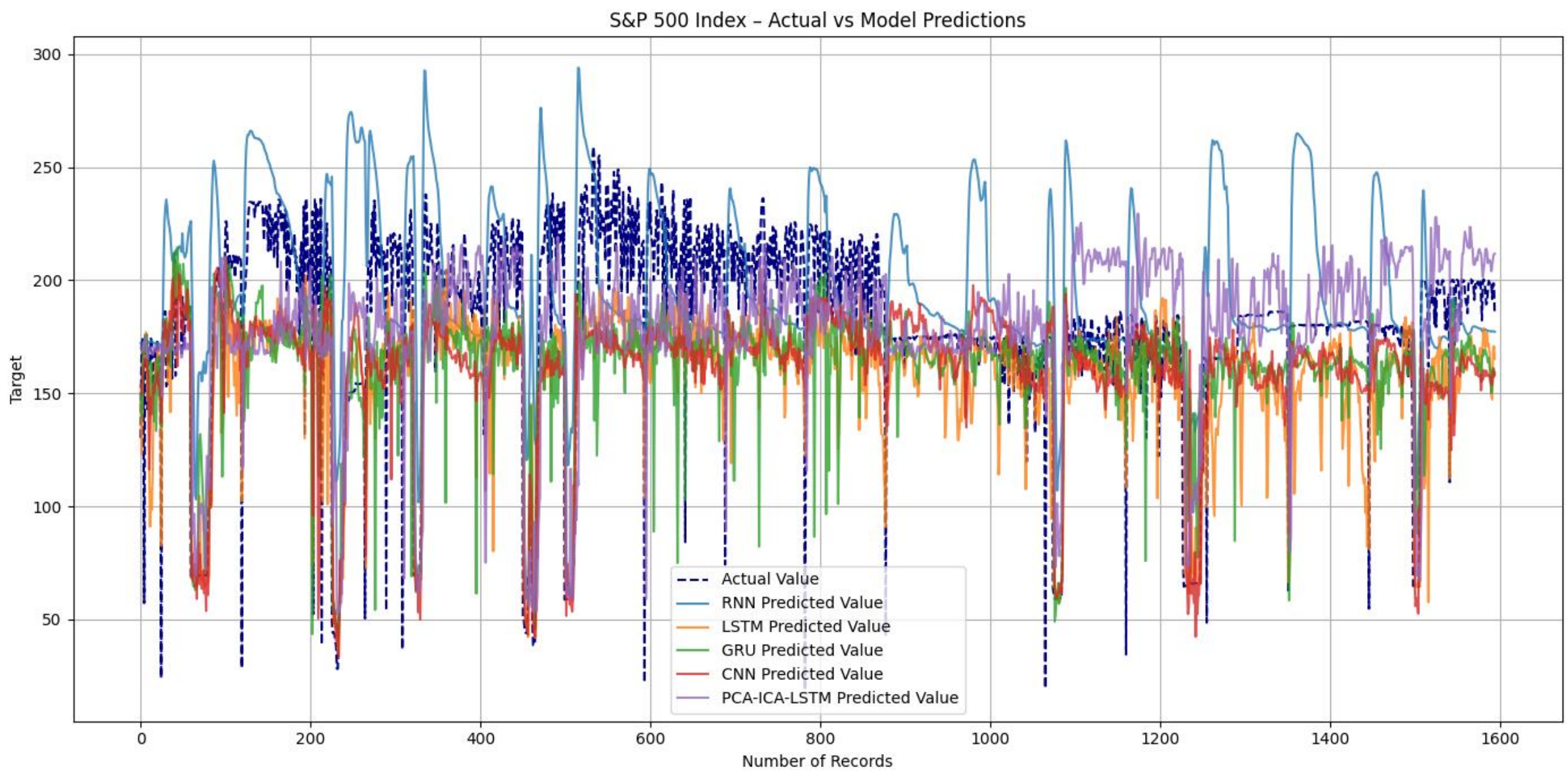

Figure 16: Actual vs. Predicted Values for S&P 500 Index Using Various Benchmark Models.

Table 9. Comparison with the benchmark models

| Model | R² | MAE | RMSE | MAPE | Flops | Inf.T (ms) | T/Ep (M) |
|---|---|---|---|---|---|---|---|
| Informer | 0.912 | 4.92 | 6.43 | 3.87% | ~3.8 | ~1.8 | ~2.6 |
| TFT-GRU | 0.883 | 5.36 | 6.88 | 4.02% | ~4.2 | ~2.1 | ~3.0 |
| GCT | 0.925 | 4.65 | 6.21 | 3.74% | ~4.8 | ~2.5 | ~3.5 |
| RNN | 0.948 | 4.31 | 5.79 | 3.51% | ~2.9 | ~1.3 | ~2.1 |
| BiLSTM | 0.946 | 4.80 | 5.92 | 3.96% | ~3.4 | ~1.5 | ~2.4 |
| CNN-LSTM | 0.952 | 4.18 | 5.63 | 3.42% | ~5.2 | ~2.8 | ~3.9 |

## 5.6 Limitations

The study's findings must be interpreted in light of key limitations [45].

- The dataset, though representative, was collected under controlled logging conditions and may not reflect the noise, missing values, and heterogeneity of real-world drilling environments, necessitating validation on diverse geological datasets.
- The models were trained in a static, supervised offline setting, limiting adaptability to

evolving operational conditions due to the absence of online or incremental learning mechanisms.

- Despite the inclusion of attention mechanisms, the hybrid architecture lacks sufficient interpretability, requiring advanced explainable AI methods to enhance transparency and operational reliability.
- Additionally, computational demands pose deployment challenges on edge devices, underscoring the need for optimization techniques such as pruning, quantization, and knowledge distillation.

# 6 Conclusions and Future Work

This study systematically investigated hybrid deep learning architectures for accurate and robust prediction of ROP in drilling operations. Starting from a baseline customized LSTM, successive enhancements, including TS-Mixer modules for nonlinear static feature fusion, attention mechanisms for adaptive weighting, and domain-tailored Transformer encoders for long-range temporal modeling, produced a unified Advanced Hybrid model. The final architecture achieved state-of-the-art predictive fidelity, with $R^2 = 0.9991$, MAE = 1.2531, RMSE = 1.6573, and MAPE = 1.1572%, surpassing all baseline and intermediate configurations. The improvements arose from complementary contributions of each component: customized LSTMs preserved sequential continuity, TS-Mixers captured complex cross-feature interactions, attention prioritized salient intervals under noisy drilling conditions, and Transformers extended contextual awareness across time. This synergy yielded enhanced adaptability to abrupt lithological transitions, minimized relative error propagation, and improved generalization across heterogeneous drilling regimes. The framework also demonstrated stable convergence with no overfitting, supported by disciplined preprocessing, regularization, and optimized training strategies. Interpretability analyses via SHAP and LIME confirmed the physical relevance of influential features like penetration time, hook load, torque, and RPM, providing transparency for operational decision-making. By integrating sequential, static, and contextual representations into a cohesive pipeline, the proposed model not only advances the methodological frontier of drilling analytics but also offers a practical foundation for real-time deployment. Its capacity to deliver precise, interpretable, and resilient ROP predictions positions it as a key enabler for drilling efficiency, proactive equipment management, and intelligent automation within petroleum engineering and broader time-series forecasting domains.

## Future Work

Future advancements in ROP prediction should emphasize model robustness, adaptability, and deployment efficiency. Expanding geologically diverse datasets with cross-well validation will enhance generalization. Adaptive learning through recursive updates, online optimization, and reinforcement strategies can enable real-time adaptation. Explainable AI methods will improve interpretability and trust. Model compression via pruning, quantization, and distillation will ensure edge deployment efficiency. Multi-task learning that jointly predicts auxiliary drilling parameters can enhance operational insight. Integrating automation, edge intelligence, and big data analytics will enable scalable, transparent, and high-performance drilling optimization.